\documentclass{article}

\PassOptionsToPackage{numbers, square}{natbib}
\usepackage[final]{nips_2018}

\usepackage[utf8]{inputenc} 
\usepackage[T1]{fontenc}    
\usepackage{hyperref}       
\usepackage{url}            
\usepackage{booktabs}       
\usepackage{amsfonts}       
\usepackage{nicefrac}       
\usepackage{microtype}      
\usepackage{amsmath}
\usepackage{amssymb}
\usepackage{graphicx}
\usepackage{titlesec}
\usepackage{caption}
\usepackage{subcaption}
\usepackage{float}
\usepackage{url}
\usepackage{todonotes}

\usepackage{caption}
\DeclareCaptionLabelFormat{cont}{#1~#2\alph{ContinuedFloat}}
\title{Improving and Understanding\\ Variational Continual Learning}

\newcommand{\KLdivK}{\mathcal{K}}
\newcommand{\KLdivL}{\mathcal{L}}
\renewcommand{\vec}[1]{\mathbf{#1}}

\author{
 Siddharth Swaroop$^{*}$, Cuong V. Nguyen$^{*}$, Thang D. Bui$^{\dagger}$, Richard E. Turner$^{*}$ \\
 $^*$ University of Cambridge, UK, \texttt{\{ss2163,vcn22,ret26\}@cam.ac.uk} \\
 $^\dagger$ University of Sydney, Australia, \texttt{thang.buivn@gmail.com}  \\
}

\begin{document}

\maketitle

\begin{abstract}
In the continual learning setting, tasks are encountered sequentially. The goal is to learn whilst i) avoiding catastrophic forgetting, ii) efficiently using model capacity, and iii) employing forward and backward transfer learning. In this paper, we explore how the Variational Continual Learning (VCL) framework achieves these desiderata on two benchmarks in continual learning: split MNIST and permuted MNIST. We first report significantly improved results on what was already a competitive approach. The improvements are achieved by establishing a new best practice approach to mean-field variational Bayesian neural networks. We then look at the solutions in detail. This allows us to obtain an understanding of why VCL performs as it does, and we compare the solution to what an `ideal' continual learning solution might be.\footnote{Code is available at https://github.com/nvcuong/variational-continual-learning/tree/master/improved\_ddm}
\end{abstract}

\section{Introduction}
\label{sec:Introduction}

In the real world, inputs can change over time, suffering from covariate and dataset shift, and often the size of datasets or privacy constraints make revisiting old data computationally prohibitive. Continual learning, where we receive data in an online fashion, allowing tasks to change and be added over time, focusses on these types of problems. An ideal continual learning solution (model and inference scheme) would, when it first sees data, efficiently use the model's capacity to solve the task, without any forgetting of contributions from old data. Once the model has reached capacity, continual learning would adjust parameters such that the `least important' information from old tasks is forgotten, and that backward and forward transfer learning is achieved: new data can be useful for increasing performance on old tasks (backward transfer), and information from old tasks can be used to perform better on new tasks (forward transfer).

We shall specifically focus on the continual image classification setting, a standard continual learning problem tackled in research. For the rest of this paper we will consider data arriving in batches as `tasks', as opposed to data arriving in a fully online or sequential manner, which is the most general form of continual learning (even though our method can be applied in this fully online setting too). Due to their success on full-batch image classification, neural networks are often employed. However, naive techniques such as vanilla neural networks suffer from catastrophic forgetting \citep{goodfellow_empirical_2013, mccloskey_catastrophic_1989}. There have been many continual learning methods introduced to combat forgetting, and we consider them to be broadly split into three groups: methods that change the model architecture, methods which use rehearsal or past data, and methods which focus on the inference scheme (see section \ref{sec:related-work}). 

 In this paper, we consider the Variational Continual Learning (VCL) framework \citep{nguyen_variational_2018}, a method which focusses on the inference scheme, but which can also be combined with a rehearsal based method by incorporating coresets. Previous inference based approaches often do not maintain accurate parameter uncertainty estimates after training on each task. Uncertainty is extremely useful in deciding the important information from previous tasks, allowing for less important information to be forgotten. A Bayesian approach to learning parameters should automatically keep uncertainty estimates, and so should perform well in the continual learning setting. We therefore consider learning a Bayesian neural network, where we set the posterior from the previous task as the prior for a new task. Exact Bayesian inference is not tractable, and so we approximate the posterior after every task. Although some of the previous inference based approaches are closely related to approximate inference in a Bayesian neural network, we optimise the variational objective function. Variational Inference (VI) is known to provide better uncertainty estimates that other approaches in this setting \citep{bui_deep_2016}.

We first summarise related work in Section \ref{sec:related-work}. We introduce the VCL algorithm in Section \ref{sec:VCL}, and report improved results on multi-head split MNIST and permuted MNIST (both with and without coresets). In Section \ref{sec:discussion}, we look in detail at the solutions we obtained. This leads to some valuable insights on the behaviour we might expect from a `good' continual learning model, both in terms of model capacity usage and forward/backward transfer, while also shedding some light on properties of mean-field variational Bayesian neural networks. Finally, we provide conclusions in Section \ref{sec:conclusions}.

\section{Related work}
\label{sec:related-work}

Here we characterise the fast-growing universe of continual learning methods in terms of three orthogonal approaches: methods that change the model architecture, methods which use rehearsal or past data, and methods which focus on the inference scheme. These approaches are complementary and so they are often combined. 

\textbf{Model based approaches to continual learning}. A conceptually simple method that changes model architecture is Progressive Neural Networks \citep{rusu_progressive_2016}, which trains a new model (neural network) for each new task, using connections between old tasks' models and newer tasks' models to incorporate forward transfer \citep{rusu_progressive_2016}. However, the overall model grows linearly in size with task, and is therefore not efficient in terms of model capacity. It also does not have potential for backward transfer. It does, however, provide extremely good results on many benchmarks. A method similar in flavour is Continual Learning via Network Pruning \citep{golkar_continual_2019}, where they now regularise during training to automatically prune out a large part of the network after every task. This method therefore handles the model capacity problem much more elegantly, but still suffers when model capacity is reached, and does not have potential for backward transfer. We shall discuss pruning more in Section \ref{sec:discussion}.

\textbf{Rehearsal and memory based approaches to continual learning}. Rehearsal of past data can be used to combat catastrophic forgetting. For example, Deep Generative Replay \citep{shin_continual_2017}, inspired by the human brain, trains a generative model on each task's data, and uses it to generate pseudo-data that is used in training in future tasks. More recently, this idea has been made more efficient by integrating the generative model into the training procedure \citep{ven_generative_2018}. Other examples of rehearsal based methods assume that a coreset of past data can be stored and used in the future \citep{nguyen_variational_2018, rebuffi_icarl_2017}. Note that storing any past data is a relaxation of the strictest continual learning problem (and may not be possible when, for example, strict privacy laws apply to past data).

\textbf{Inference based approaches to continual learning}. The third set of approaches regularise parameter updates by penalising against changes in `important' parameters for previous tasks. For example, Elastic Weight Consolidation (EWC) \citep{kirkpatrick_overcoming_2017, ritter_online_2018} regularises using Fisher information matrices from previous tasks. Synaptic Intelligence (SI) \citep{zenke_continual_2017} compares the (rate of change of) the objective's gradient and the parameters to decide each parameter's importance to previous tasks, and EWC and SI can be combined \citep{chaudhry_riemannian_2018}. Variational continual learning without coresets is also a pure inference based approach to continual learning.

\textbf{Mixed approaches}. It seems likely that using some combination of the three approaches to continual learning will provide the best methods. Indeed, VCL with coresets is such a mixed approach. Similarly, Progress \& Compress \citep{schwarz_progress_2018} use concepts from Progressive Neural Networks \citep{rusu_progressive_2016} and EWC \citep{kirkpatrick_overcoming_2017}. They train an active column on each new task (with some layerwise connections to the `knowledge base'), which is then distilled into a knowledge base, using a modified version of EWC to prevent catastrophic forgetting.

\section{Improvements to Variational Continual Learning}
\label{sec:VCL}

VCL adopts a Bayesian approach to learning parameters of a model, using the previous task's posterior as the new task's prior when we see new data. VCL approximates each task's posterior by a variational distribution, $q_t(\vec{\theta}) \approx p_t(\vec{\theta | \mathcal{D}_{1:t}})$, obtained by using Monte Carlo variational inference \citep{blundell_weight_2015}, maximising the objective (for every task $t$)
\begin{equation} \label{eq:VI function}
    \mathcal{L}_t(q_t(\vec{\theta})) = \sum_{n=1}^{N_t} \mathbb{E}_{\vec{\theta} \sim q_t(\vec{\theta})} \left[ \log{p(y_t^{(n)}|\vec{\theta},\vec{x}_t^{(n)}}) \right] - \KLdivK \KLdivL \left( q_t(\vec{\theta}) || q_{t-1}(\vec{\theta}) \right).
\end{equation}
There are two terms in Equation \ref{eq:VI function}, which we shall refer to as the likelihood term (the left hand term with Monte Carlo estimates), and the $\KLdivK \KLdivL$ term. For further details on the optimisation process, please see the original VCL paper \citep{nguyen_variational_2018}. We consider fully-connected multi-layer perceptrons in this paper, with ReLU activation functions, and use a Gaussian mean-field approximation over each of the weights, as in the original paper. We place a standard normal Gaussian over each parameter as the initial prior ($p_0(\vec{\theta})$), and can calculate the $\KLdivK \KLdivL$ term in Equation \ref{eq:VI function} analytically. We use ADAM \citep{kingma_adam_2014} to optimise for the means and variances of each parameter. For tests with episodic memory enhancement (coresets), we treat coresets the same way as in \citet{nguyen_variational_2018}, and randomly choose the examples in the coreset (the `random coreset' method). We remove the (randomly selected) coreset images while training on the rest of the data, and train on only the coreset images just before test-time.

\subsection{Improved results}

We established a simple scheme for substantially improving results over \citet{nguyen_variational_2018}. Crucially, we optimise for a much longer time. Although progress can sometimes appear to stall during optimisation, in reality progress is just extremely slow. We also firstly introduce the local reparameterisation trick during the Monte Carlo sampling of the network \citep{kingma_variational_2015}, and secondly initialise all weights with small, random means (of the order $10^{-1}$) and small variances (of the order $10^{-3}$) before every task. Without these additional two tricks, we would have to wait for much longer in order to reach convergence. Although these improvements may seem technically trivial, we find that they are practically very important, greatly impacting results.

In the multi-head split MNIST task, we have to sequentially solve five binary classification tasks from the MNIST dataset: \{0v1\}, \{2v3\}, \{4v5\}, \{6v7\}, \{8v9\}. The challenge in split MNIST is to obtain good performance on new tasks while retaining performance on old ones. We ran a one hidden layer model with 200 units for 600 epochs (with 256 batch size), sharing the lower level weights between tasks, and report the mean and standard deviation over 10 runs. See Table \ref{table:split mnist} for results. Without coresets, we achieve a final test accuracy of 98.5$\pm$0.4\%, an increase from 97.0\% reported earlier \citep{nguyen_variational_2018}. With coresets, we achieve 98.2$\pm$0.4\%, similar to the 98.4\% reported earlier.

We also consider permuted MNIST. In permuted MNIST, we receive tasks sequentially, where each task is the standard (10-way) MNIST classification task, with the pixels having undergone a fixed random permutation, different for each task. Ideally, a network with two or more hidden layers would use lower layer(s) to `de-permute' the images, and higher layer(s) to solve MNIST, which is then constant between tasks. We ran a two hidden layer model with 100 units in each hidden layer for 800 epochs (with 1024 batch size), and report the mean and standard deviation over 5 runs. Without coresets, this achieves a final average test accuracy of 93$\pm$1\%, as compared to 90\% reported earlier \citep{nguyen_variational_2018} (Table \ref{table:permuted mnist}). With coresets, we achieve a final average test accuracy of 94.6$\pm$0.3\%, as compared to 93\% reported earlier. For reference, training the same network but seeing all the data together (batch mode) has an accuracy of 97\% (this is an upper bound on the performance possible with this model and inference scheme).

\begin{table}
\caption{Final average test accuracy on multi-head split MNIST. VCL: Variational Continual Learning, EWC: Elastic Weight Consolidation, SI: Synaptic Intelligence. Results for EWC and SI are taken from \citet{nguyen_variational_2018}.}
\centering
\label{table:split mnist}
\begin{tabular}{l l p{.2\textwidth}}
\toprule
Method & Split MNIST & Split MNIST \par + 40 random coreset \\
\midrule
VCL (this paper)                                & 98.5$\pm$0.4\%    & 98.2$\pm$0.4\%\\
Previous VCL \citep{nguyen_variational_2018}    & 97.0\%            & 98.4\%        \\
EWC \citep{kirkpatrick_overcoming_2017}         & 63.1\%            & -             \\
SI \citep{zenke_continual_2017}                 & 98.9\%            & -             \\
\bottomrule
\end{tabular}
\end{table}

\section{Discussion}
\label{sec:discussion}

We now look into how VCL continually learns tasks, exploring how it uses its model capacity, primarily by looking at the learnt weights in the network.

\subsection{Split MNIST}

\begin{figure}[h]
\centering
    \includegraphics[scale=0.46]{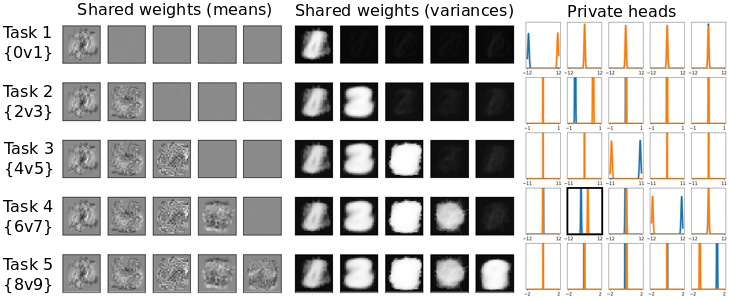}
    \caption{The active units learnt during split MNIST, without coresets. Rows correspond to stages of continual learning. Left: means, Centre: variances, Right: output weights for each task's two classes. Exactly the same effect is observed when incorporating coresets.}
    \label{fig:split MNIST neurons of interest}
\end{figure}

We first consider the model trained without coresets (although exactly the same effects happen with coresets). Appendix \ref{app:split} shows plots of weights into and out of each unit after training on each task. Even though we have 200 units in the single hidden layer, it appears as if only one unit is being used per each of the five tasks; the remaining units are pruned out as part of the optimisation process. These active (un-pruned) units are plotted in Figure \ref{fig:split MNIST neurons of interest}. This effect is similar to that observed in \citet{trippe_overpruning_2018}, with entire units pruned out, as opposed to just individual weights. The pruned units appear to have input weights at or near the prior (standard normal Gaussian), with output weights near a delta function (zero mean, small variance), therefore minimising their effect on the output prediction. Removing all pruned units from the network does not change the network's predictions (and therefore accuracy).

This pruning effect seems to be due to the choice of inference scheme. We leave a detailed review of the pruning process for future work, including characterising the mechanism and explaining it mathematically. Intuitively, the pruning effect can be explained by looking at the optimisation function (Equation \ref{eq:VI function}). By reducing the effect of a unit on the output prediction (setting output weights to have zero mean and small variance), the input weights to the unit can be set to their prior. The increase in the $\KLdivK \KLdivL$ term due to the small variance of the output weights are offset by the reduction in the $\KLdivK \KLdivL$ term from the numerically more input weights. Provided the likelihood term does not change too much, a pruned solution is therefore more optimal. In this paper, we now focus on what our pruned solutions reveal about how our model approaches continual learning tasks, and debate whether the pruning effect is a feature or a bug for continual learning.

As the model only uses 1 hidden unit per task, it has learnt to use a fraction of its total capacity to successfully learn binary classifications in the split MNIST experiment. The high test accuracies indicate this is an efficient use of model capacity. Remaining, unused units can be used for other tasks that we may see in the future. This implies that pruning is a beneficial feature for continual learning, forcing the model to efficiently use its capacity.

Additionally, the pruning effect allows us to see some forward and backward transfer (see Figure \ref{fig:split MNIST neurons of interest}), both important qualities in a good continual learning solution. Forward transfer is visible when previous tasks' active units have non-zero weights for subsequent tasks. For example, unit 2, which was learnt after task 2 (classifying digits \{2v3\}), has non-zero output weights after task 4 (classifying \{6v7\}). The model therefore uses some information about the task \{2v3\} in solving the task \{6v7\}. Although less visible in the plots, there is also backward transfer in the same units: unit 2's input weights change slightly after training on task 4 (\{6v7\}), potentially changing test accuracy on task 2. In this case however, any backward transfer does not result in different test accuracies; this could be because there is no potential for improvement given the high test accuracies involved.

The exact same effects are seen when we train the model with coresets: the same pruning effect appears (Figure \ref{fig:split MNIST neurons of interest} and the figures in Appendix \ref{app:split} are similar when trained with coresets), and similar test accuracies are obtained (within one standard deviation). This shows that for a simple task such as multi-head split MNIST, there is no need for incorporating coresets.

This pruning effect is also similar to that considered in \citet{golkar_continual_2019}, where they prune out entire units. However, they have hyperparameters that control the degree of pruning (and corresponding accuracy loss). Previous work on Variational Inference for BNNs has found that VI methods can be used to prune large parts of the network \citep{louizos_bayesian_2017}, and we show here how this pruning is done over units as opposed to weights (see also \citet{trippe_overpruning_2018}). In comparison to \citet{golkar_continual_2019}, our method automatically prunes out entire units, and is able to re-use the units for both forward and backward transfer.

%
%
\subsection{Permuted MNIST}

\begin{figure}[h]
\centering
    \includegraphics[scale=0.46]{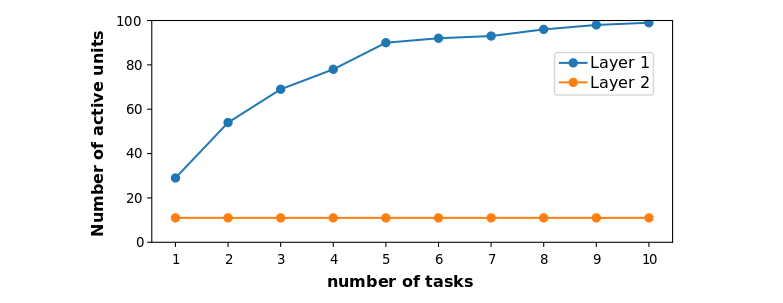}
    \caption{Number of active units per hidden layer after each task in permuted MNIST, without coresets. Exactly the same effect is observed when incorporating coresets.}
    \label{fig:un-pruned units permuted}
\end{figure}

We now look into how the two hidden layer model approaches permuted MNIST (results in Table \ref{table:permuted mnist}). We first consider the model trained without coresets. Appendix \ref{app:permuted} shows figures of the weights, showing there is still pruning. The numbers of active (un-pruned) units after training on each task are summarised in Figure \ref{fig:un-pruned units permuted}.

There are more active units in permuted MNIST than were in split MNIST, perhaps due to the more difficult nature of permuted MNIST (classifying between 10 digits, as opposed to between 2). However, only 11 units are used in the second hidden layer, with the remaining 89 units pruned out. Additionally, the output weights on these 11 units do not change between tasks (see Appendix \ref{app:permuted} for weights for training without coresets; the figures with coresets look the same). This confirms that the hidden layers effectively de-permute the images, allowing the output weights to just classify between the 10 digits.

Beyond re-using the upper level weights, there is not much evidence of forward or backward transfer. We should expect this from permuted MNIST because the network trains on all MNIST digits on the first task itself, hence already learning the `best' way to classify between MNIST digits. Any subsequent permuted images cannot improve this. Instead, the remaining focus of permuted MNIST seems to be on ensuring we use available model capacity as efficiently as possible. Increasing the model capacity improves results: training a network with 250 units in the lower hidden layer (instead of 100) improves final average test accuracy to 95.5\% (10 tasks).

Incorporating coresets also improves results. However, the number of active units (plotted in Figures \ref{fig:un-pruned units permuted}) and the weight plots (in Appendix \ref{app:permuted}) look the same. Instead, training VCL with coresets (which can be viewed as changing the order in which the model trains on data, or, changing the schedule with which we visit training data) appears to reinforce previous tasks' images: it lowers forgetting in the network.

\begin{table}
\caption{Final average test accuracy on permuted MNIST for various methods (results taken from respective papers). A hidden layer size of \{$n_1, n_2$\} indicates two hidden layers, the lower hidden layer having $n_1$ hidden units, the upper hidden layer having $n_2$ (followed by a softmax over the 10 MNIST classes). Methods with an asterisk (*) use some sort of episodic memory. Results marked with a double asterisk (**) were read from a graph in the source paper. }
\centering
\label{table:permuted mnist}
\begin{tabular}{l p{.14\textwidth} p{.1\textwidth} p{.2\textwidth}}
\toprule
Method/Paper & Hidden \par layer size & Number \par of tasks & Final average \par test accuracy \\
\midrule
VCL (this paper)                                                                                & \{100, 100\}      & 10    & 93$\pm$1\%        \\
*VCL + 200 random coreset (this paper)                                                          & \{100, 100\}      & 10    & 94.6$\pm$0.3\%    \\
Previous VCL \citep{nguyen_variational_2018}                                                    & \{100, 100\}      & 10    & 90\%              \\
*Previous VCL + 200 random coreset \citep{nguyen_variational_2018}                              & \{100, 100\}      & 10    & 93\%              \\
Kronecker-factored Laplace \citep{ritter_online_2018}                                           & \{100, 100\}      & 50    & 90\%              \\
Kronecker-factored Laplace \citep{ritter_online_2018}                                           & \{100, 100\}      & 10    & 96\%**            \\
EWC \citep{kirkpatrick_overcoming_2017}                                                         & \{2000, 2000\}    & 10    & 97\%              \\
EWC \citep{kirkpatrick_overcoming_2017} (result from \citep{nguyen_variational_2018})           & \{100, 100\}      & 10    & 84\%              \\
SI \citep{zenke_continual_2017}                                                                 & \{2000, 2000\}    & 10    & 97\%              \\
SI \citep{zenke_continual_2017} (result from \citep{nguyen_variational_2018})                   & \{100, 100\}      & 10    & 86\%              \\
Continual Learning via Neural Pruning \citep{golkar_continual_2019}                             & \{2000, 2000\}    & 10    & 98.42$\pm$0.04\%  \\
*A-GEM \citep{chaudhry_efficient_2018}                                                          & \{256, 256\}      & 20    & 89.1$\pm$0.14\%   \\
*A-GEM \citep{chaudhry_efficient_2018}                                                          & \{256, 256\}      & 10    & 92.3\%**          \\
*GEM \citep{lopez_gradient_2017}                                                                & \{100, 100\}      & 20    & 80\%              \\
BLLL-REG \citep{ebrahimi_uncertainty_2019}                                                      
& \{100, 100\}      & 10    & 92.2\%            \\
*GEM \citep{lopez_gradient_2017} (result from \citep{chaudhry_efficient_2018})                  & \{256, 256\}      & 20    & 89.5$\pm$0.48\%   \\
*GEM \citep{lopez_gradient_2017} (result from \citep{chaudhry_efficient_2018})                  & \{256, 256\}      & 10    & 93.1\%**          \\
Riemannian Walk \citep{chaudhry_riemannian_2018} (result from \citep{chaudhry_efficient_2018})  & \{256, 256\}      & 20    & 85.7$\pm$0.56\%   \\
Riemannian Walk \citep{chaudhry_riemannian_2018} (result from \citep{chaudhry_efficient_2018})  & \{256, 256\}      & 10    & 91.6\%**          \\
Progressive NNs \citep{rusu_progressive_2016} (result from \citep{chaudhry_efficient_2018})     & \{256, 256\}      & 20    & 93.5$\pm$0.07\%   \\
Progressive NNs \citep{rusu_progressive_2016} (result from \citep{chaudhry_efficient_2018})     & \{256, 256\}      & 10    & 94.6\%**          \\
\bottomrule
\end{tabular}
\end{table}

Table \ref{table:permuted mnist} summarises some recent works' results on permuted MNIST. As can be seen, different papers use different numbers of hidden units in their hidden layers, and test over different numbers of tasks. However, as the findings in this paper indicate, permuted MNIST primarily tests for model capacity, highlighting how we cannot faithfully compare results when model capacity and number of tasks differ. We propose that when using permuted MNIST as a benchmark in continual learning, model capacity is kept fairly limited (for example, two hidden layers with either 100 or 256 hidden units each), and number of tasks is kept high (minimum 10, possibly more). Of all methods, \citet{ritter_online_2018} achieves significantly superior results, as they use a relatively small network and test over many tasks. Note that 2000 units in two hidden layers is an extremely large model for 10 tasks: as \citet{golkar_continual_2019} note, they can achieve high (as good as single-task) performance having pruned out a large proportion of their network. With such a large model capacity, this benchmark no longer tests any desiderata in continual learning aside from avoiding catastrophic forgetting, which many other benchmarks can also do.

Many papers often use permuted MNIST to demonstrate how their method (and other baseline methods) exhibit some of continual learning's desiderata. For example, they show some form of resistance to forgetting or efficient use of model capacity via metrics or plots. In this case, comparing the final average test accuracy is no longer so important, but we believe far better comparisons would be achieved if a more challenging hidden layer size and number of tasks was used.

\section{Conclusions and future work}
\label{sec:conclusions}

This paper has significantly improved the performance of Variational Continual Learning  \citep{nguyen_variational_2018} on split MNIST and permuted MNIST. It has also shed light on the mechanism that VCL employs to counter catastrophic forgetting. Future work would involve applying VCL to other, harder benchmarks, where there is more potential for forward and backward transfer. It should also explore the observed pruning effect, a consequence of using mean-field variational Bayesian neural networks. Another avenue for future work would be to consider structured approximations to the distribution over parameters (we currently only consider the mean-field approximation), which can potentially improve results in continual learning \citep{ritter_online_2018}.

%
%
\newpage
\small

\bibliographystyle{abbrvnat}
\bibliography{references}

\begin{thebibliography}{22}
\providecommand{\natexlab}[1]{#1}
\providecommand{\url}[1]{\texttt{#1}}
\expandafter\ifx\csname urlstyle\endcsname\relax
  \providecommand{\doi}[1]{doi: #1}\else
  \providecommand{\doi}{doi: \begingroup \urlstyle{rm}\Url}\fi

\bibitem[Blundell et~al.(2015)Blundell, Cornebise, Kavukcuoglu, and
  Wierstra]{blundell_weight_2015}
C.~Blundell, J.~Cornebise, K.~Kavukcuoglu, and D.~Wierstra.
\newblock Weight uncertainty in neural network.
\newblock In F.~Bach and D.~Blei, editors, \emph{Proceedings of the 32nd
  International Conference on Machine Learning}, volume~37 of \emph{Proceedings
  of Machine Learning Research}, pages 1613--1622, Lille, France, 07--09 Jul
  2015. PMLR.
\newblock URL \url{http://proceedings.mlr.press/v37/blundell15.html}.

\bibitem[Bui et~al.(2016)Bui, Hernandez-Lobato, Hernandez-Lobato, Li, and
  Turner]{bui_deep_2016}
T.~Bui, D.~Hernandez-Lobato, J.~Hernandez-Lobato, Y.~Li, and R.~Turner.
\newblock Deep gaussian processes for regression using approximate expectation
  propagation.
\newblock In M.~F. Balcan and K.~Q. Weinberger, editors, \emph{Proceedings of
  The 33rd International Conference on Machine Learning}, volume~48 of
  \emph{Proceedings of Machine Learning Research}, pages 1472--1481, New York,
  New York, USA, 20--22 Jun 2016. PMLR.
\newblock URL \url{http://proceedings.mlr.press/v48/bui16.html}.

\bibitem[Chaudhry et~al.(2018)Chaudhry, Dokania, Ajanthan, and
  Torr]{chaudhry_riemannian_2018}
A.~Chaudhry, P.~K. Dokania, T.~Ajanthan, and P.~H.~S. Torr.
\newblock Riemannian walk for incremental learning: Understanding forgetting
  and intransigence.
\newblock \emph{CoRR}, abs/1801.10112, 2018.
\newblock URL \url{http://arxiv.org/abs/1801.10112}.

\bibitem[Chaudhry et~al.(2019)Chaudhry, Ranzato, Rohrbach, and
  Elhoseiny]{chaudhry_efficient_2018}
A.~Chaudhry, M.~Ranzato, M.~Rohrbach, and M.~Elhoseiny.
\newblock Efficient lifelong learning with a-{GEM}.
\newblock In \emph{International Conference on Learning Representations}, 2019.
\newblock URL \url{https://openreview.net/forum?id=Hkf2_sC5FX}.

\bibitem[Ebrahimi et~al.(2019)Ebrahimi, Elhoseiny, Darrell, and
  Rohrbach]{ebrahimi_uncertainty_2019}
S.~Ebrahimi, M.~Elhoseiny, T.~Darrell, and M.~Rohrbach.
\newblock Uncertainty-guided lifelong learning in bayesian networks, 2019.
\newblock URL \url{https://openreview.net/forum?id=SJMBM2RqKQ}.

\bibitem[Golkar et~al.(2019)Golkar, Kagan, and Cho]{golkar_continual_2019}
S.~Golkar, M.~Kagan, and K.~Cho.
\newblock Continual learning via neural pruning.
\newblock \emph{CoRR}, abs/1903.04476, 2019.
\newblock URL \url{http://arxiv.org/abs/1903.04476}.

\bibitem[Goodfellow et~al.(2014)Goodfellow, Mirza, Xiao, Courville, and
  Bengio]{goodfellow_empirical_2013}
I.~J. Goodfellow, M.~Mirza, D.~Xiao, A.~Courville, and Y.~Bengio.
\newblock An {Empirical} {Investigation} of {Catastrophic} {Forgetting} in
  {Gradient}-{Based} {Neural} {Networks}.
\newblock In \emph{International Conference on Learning Representations}, 2014.
\newblock URL \url{http://arxiv.org/abs/1312.6211}.

\bibitem[Kingma and Ba(2014)]{kingma_adam_2014}
D.~P. Kingma and J.~Ba.
\newblock Adam: {A} method for stochastic optimization.
\newblock \emph{CoRR}, abs/1412.6980, 2014.
\newblock URL \url{http://arxiv.org/abs/1412.6980}.

\bibitem[Kingma et~al.(2015)Kingma, Salimans, and
  Welling]{kingma_variational_2015}
D.~P. Kingma, T.~Salimans, and M.~Welling.
\newblock Variational {Dropout} and the {Local} {Reparameterization} {Trick}.
\newblock In C.~Cortes, N.~D. Lawrence, D.~D. Lee, M.~Sugiyama, and R.~Garnett,
  editors, \emph{Advances in {Neural} {Information} {Processing} {Systems} 28},
  pages 2575--2583. Curran Associates, Inc., 2015.
\newblock URL
  \url{http://papers.nips.cc/paper/5666-variational-dropout-and-the-local-reparameterization-trick.pdf}.

\bibitem[Kirkpatrick et~al.(2017)Kirkpatrick, Pascanu, Rabinowitz, Veness,
  Desjardins, Rusu, Milan, Quan, Ramalho, Grabska-Barwinska, Hassabis, Clopath,
  Kumaran, and Hadsell]{kirkpatrick_overcoming_2017}
J.~Kirkpatrick, R.~Pascanu, N.~Rabinowitz, J.~Veness, G.~Desjardins, A.~A.
  Rusu, K.~Milan, J.~Quan, T.~Ramalho, A.~Grabska-Barwinska, D.~Hassabis,
  C.~Clopath, D.~Kumaran, and R.~Hadsell.
\newblock Overcoming catastrophic forgetting in neural networks.
\newblock \emph{Proceedings of the National Academy of Sciences, 2017}, 2017.
\newblock URL \url{http://arxiv.org/abs/1612.00796}.
\newblock arXiv: 1612.00796.

\bibitem[Lopez-Paz and Ranzato(2017)]{lopez_gradient_2017}
D.~Lopez-Paz and M.~A. Ranzato.
\newblock Gradient episodic memory for continual learning.
\newblock In I.~Guyon, U.~V. Luxburg, S.~Bengio, H.~Wallach, R.~Fergus,
  S.~Vishwanathan, and R.~Garnett, editors, \emph{Advances in Neural
  Information Processing Systems 30}, pages 6467--6476. Curran Associates,
  Inc., 2017.
\newblock URL
  \url{http://papers.nips.cc/paper/7225-gradient-episodic-memory-for-continual-learning.pdf}.

\bibitem[Louizos et~al.(2017)Louizos, Ullrich, and
  Welling]{louizos_bayesian_2017}
C.~Louizos, K.~Ullrich, and M.~Welling.
\newblock Bayesian compression for deep learning.
\newblock In I.~Guyon, U.~V. Luxburg, S.~Bengio, H.~Wallach, R.~Fergus,
  S.~Vishwanathan, and R.~Garnett, editors, \emph{Advances in Neural
  Information Processing Systems 30}, pages 3288--3298. Curran Associates,
  Inc., 2017.
\newblock URL
  \url{http://papers.nips.cc/paper/6921-bayesian-compression-for-deep-learning.pdf}.

\bibitem[Mccloskey and Cohen(1989)]{mccloskey_catastrophic_1989}
M.~Mccloskey and N.~J. Cohen.
\newblock Catastrophic interference in connectionist networks: {T}he sequential
  learning problem.
\newblock \emph{The Psychology of Learning and Motivation}, 24:\penalty0
  104--169, 1989.

\bibitem[Nguyen et~al.(2018)Nguyen, Li, Bui, and
  Turner]{nguyen_variational_2018}
C.~V. Nguyen, Y.~Li, T.~D. Bui, and R.~E. Turner.
\newblock Variational continual learning.
\newblock In \emph{International Conference on Learning Representations}, 2018.
\newblock URL \url{https://openreview.net/forum?id=BkQqq0gRb}.

\bibitem[Rebuffi et~al.(2017)Rebuffi, Kolesnikov, Sperl, and
  Lampert]{rebuffi_icarl_2017}
S.~Rebuffi, A.~Kolesnikov, G.~Sperl, and C.~H. Lampert.
\newblock icarl: Incremental classifier and representation learning.
\newblock In \emph{2017 {IEEE} Conference on Computer Vision and Pattern
  Recognition, {CVPR} 2017, Honolulu, HI, USA, July 21-26, 2017}, pages
  5533--5542, 2017.
\newblock \doi{10.1109/CVPR.2017.587}.
\newblock URL \url{https://doi.org/10.1109/CVPR.2017.587}.

\bibitem[Ritter et~al.(2018)Ritter, Botev, and Barber]{ritter_online_2018}
H.~Ritter, A.~Botev, and D.~Barber.
\newblock Online structured laplace approximations for overcoming catastrophic
  forgetting.
\newblock In S.~Bengio, H.~Wallach, H.~Larochelle, K.~Grauman, N.~Cesa-Bianchi,
  and R.~Garnett, editors, \emph{Advances in Neural Information Processing
  Systems 31}, pages 3738--3748. Curran Associates, Inc., 2018.
\newblock URL
  \url{http://papers.nips.cc/paper/7631-online-structured-laplace-approximations-for-overcoming-catastrophic-forgetting.pdf}.

\bibitem[Rusu et~al.(2016)Rusu, Rabinowitz, Desjardins, Soyer, Kirkpatrick,
  Kavukcuoglu, Pascanu, and Hadsell]{rusu_progressive_2016}
A.~A. Rusu, N.~C. Rabinowitz, G.~Desjardins, H.~Soyer, J.~Kirkpatrick,
  K.~Kavukcuoglu, R.~Pascanu, and R.~Hadsell.
\newblock Progressive neural networks.
\newblock \emph{CoRR}, abs/1606.04671, 2016.
\newblock URL \url{http://arxiv.org/abs/1606.04671}.

\bibitem[Schwarz et~al.(2018)Schwarz, Luketina, Czarnecki, Grabska-Barwinska,
  Teh, Pascanu, and Hadsell]{schwarz_progress_2018}
J.~Schwarz, J.~Luketina, W.~M. Czarnecki, A.~Grabska-Barwinska, Y.~W. Teh,
  R.~Pascanu, and R.~Hadsell.
\newblock Progress \& {Compress}: {A} scalable framework for continual
  learning.
\newblock \emph{ICML 2018}, May 2018.
\newblock URL \url{http://arxiv.org/abs/1805.06370}.
\newblock arXiv: 1805.06370.

\bibitem[Shin et~al.(2017)Shin, Lee, Kim, and Kim]{shin_continual_2017}
H.~Shin, J.~K. Lee, J.~Kim, and J.~Kim.
\newblock Continual learning with deep generative replay.
\newblock In I.~Guyon, U.~V. Luxburg, S.~Bengio, H.~Wallach, R.~Fergus,
  S.~Vishwanathan, and R.~Garnett, editors, \emph{Advances in Neural
  Information Processing Systems 30}, pages 2990--2999. Curran Associates,
  Inc., 2017.
\newblock URL
  \url{http://papers.nips.cc/paper/6892-continual-learning-with-deep-generative-replay.pdf}.

\bibitem[Trippe and Turner(2018)]{trippe_overpruning_2018}
B.~Trippe and R.~Turner.
\newblock Overpruning in {Variational} {Bayesian} {Neural} {Networks}.
\newblock \emph{arXiv:1801.06230 [stat]}, Jan. 2018.
\newblock URL \url{http://arxiv.org/abs/1801.06230}.
\newblock arXiv: 1801.06230.

\bibitem[van~der Ven and Tolias(2018)]{ven_generative_2018}
G.~M. van~der Ven and A.~S. Tolias.
\newblock Generative replay with feedback connections as a general strategy for
  continual learning.
\newblock \emph{CoRR}, abs/1809.10635, 2018.
\newblock URL \url{http://arxiv.org/abs/1809.10635}.

\bibitem[Zenke et~al.(2017)Zenke, Poole, and Ganguli]{zenke_continual_2017}
F.~Zenke, B.~Poole, and S.~Ganguli.
\newblock Continual learning through synaptic intelligence.
\newblock In D.~Precup and Y.~W. Teh, editors, \emph{Proceedings of the 34th
  International Conference on Machine Learning}, volume~70 of \emph{Proceedings
  of Machine Learning Research}, pages 3987--3995, International Convention
  Centre, Sydney, Australia, 06--11 Aug 2017. PMLR.
\newblock URL \url{http://proceedings.mlr.press/v70/zenke17a.html}.

\end{thebibliography}
\newpage
\titleformat{\section}{\large\bfseries}{\appendixname~\thesection}{0.5em}{}
\appendix

\section{} \label{app:accuracy plots}

This appendix has figures showing the test accuracy after training on split MNIST and permuted MNIST, both with and without coresets.

\begin{figure}[H]
\centering
    \includegraphics[scale=0.25]{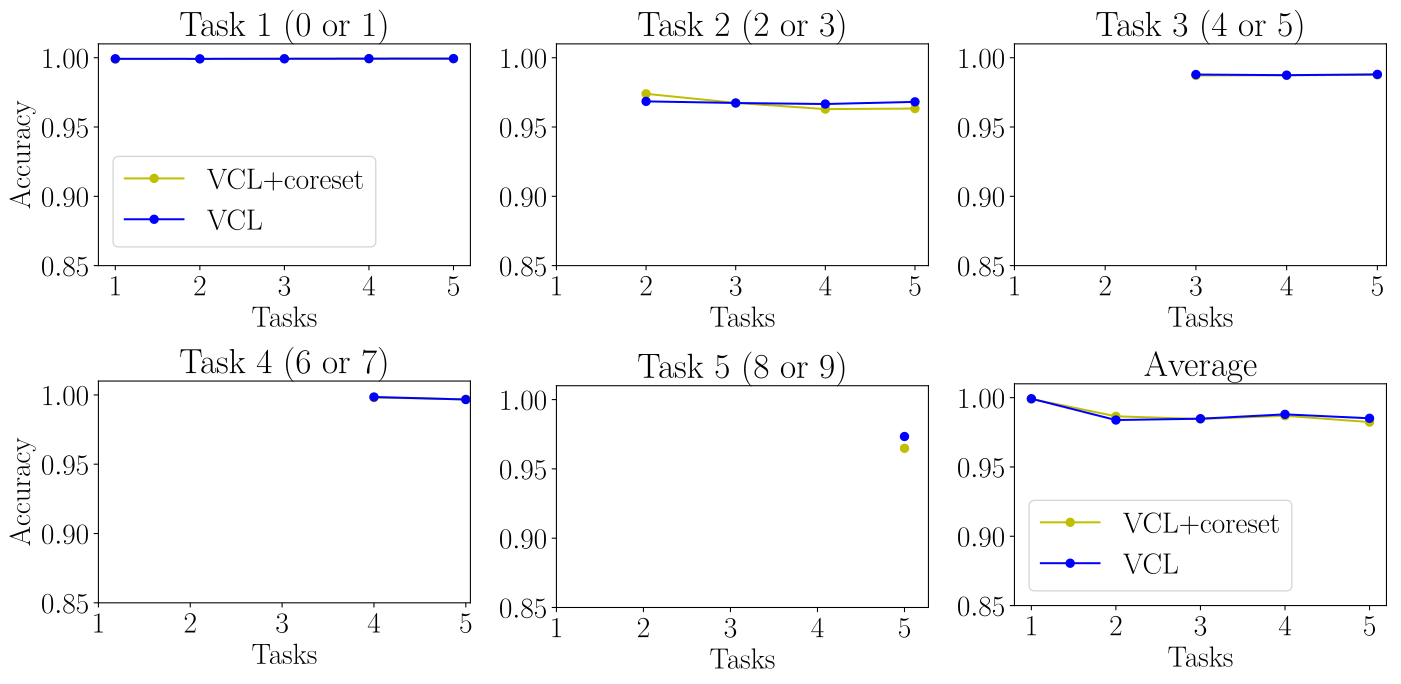}
    \caption{Test accuracy on all tasks for split MNIST, averaged over 10 runs. The bottom-left plot shows the average accuracy over all tasks. The coreset used was 40 randomly chosen examples from each task.}
    \label{fig:split mnist accuracy}
\end{figure}

\begin{figure}[H]
\centering
    \includegraphics[scale=0.5]{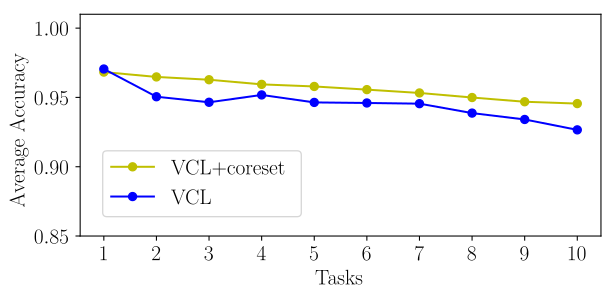}
    \caption{Average test accuracy after each task for permuted MNIST, averaged over 5 runs. The coreset used was 200 randomly chosen examples from each task.}
    \label{fig:permuted mnist accuracy}
\end{figure}

\section{} \label{app:split}
This appendix has figures showing the means and variances of weights after training without coresets on each of the five tasks in split MNIST (MLP with 1 hidden layers, 200 hidden units).

\begin{figure}[H]
\centering
    \ContinuedFloat*
    \includegraphics[scale=0.25]{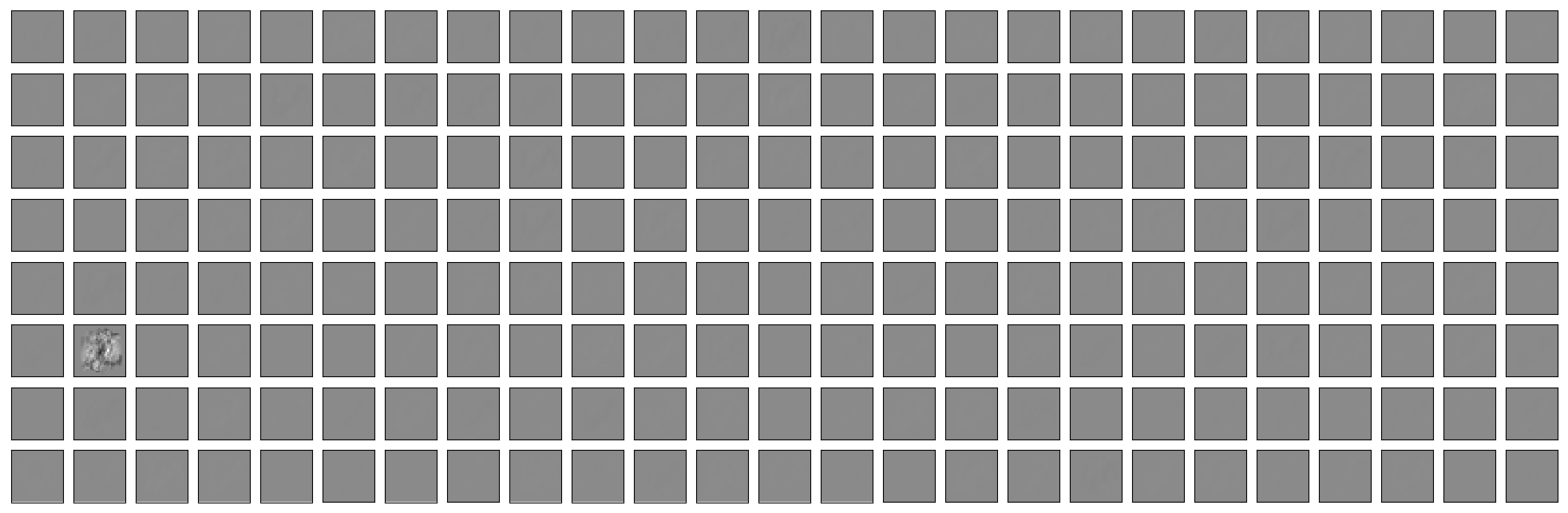}
    \caption{Hidden layer input means after training on task 1}
    \label{fig:split task1 layer0 mean}
\end{figure}

\begin{figure}[H]
\centering
    \ContinuedFloat
    \includegraphics[scale=0.25]{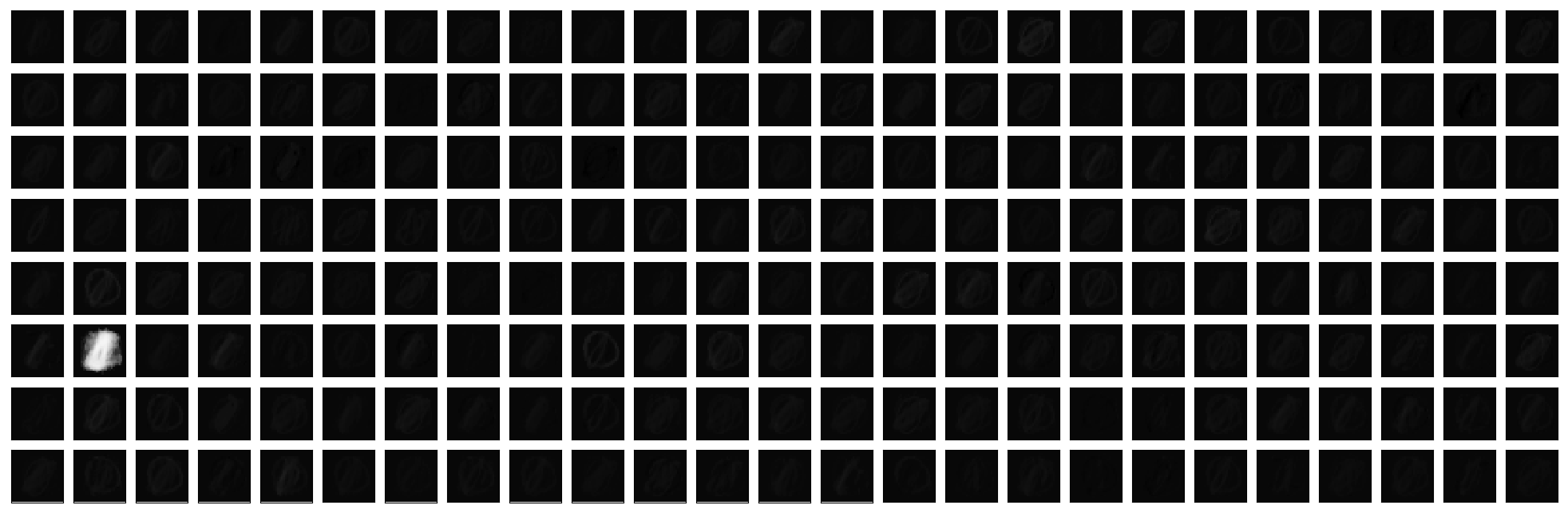}
    \caption{Hidden layer input variances after training on task 1}
    \label{fig:split task1 layer0 var}
\end{figure}

\begin{figure}[H]
\centering
    \ContinuedFloat
    \includegraphics[scale=0.25]{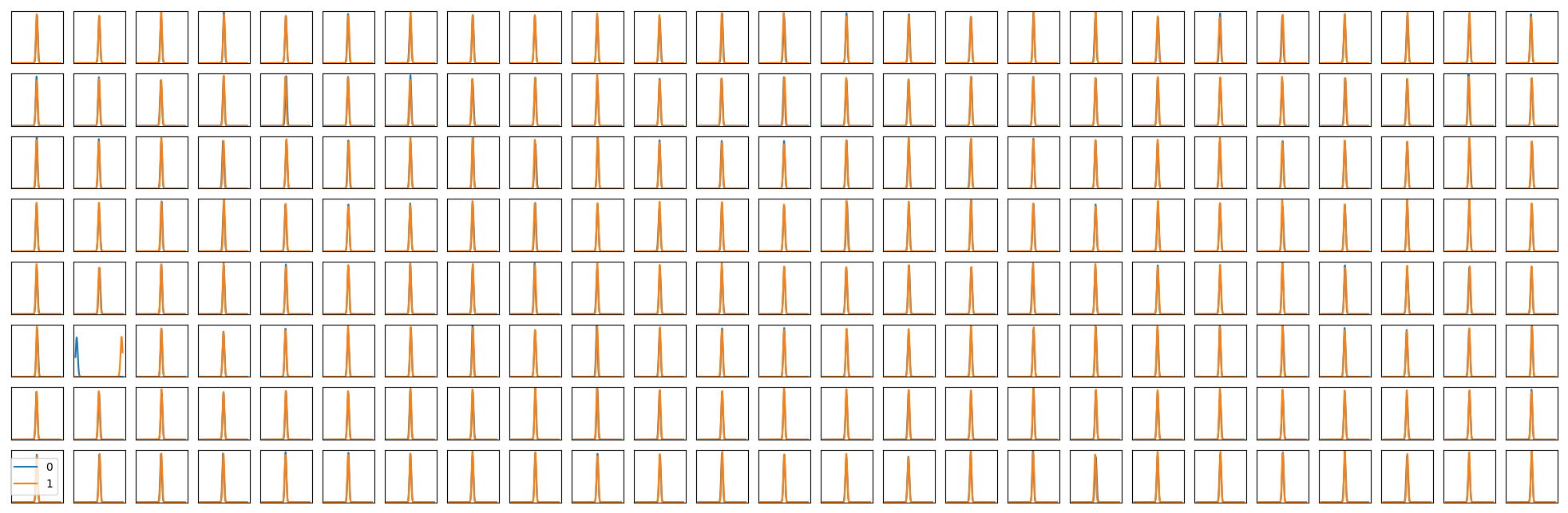}
    \caption{Hidden layer output weights after training on task 1}
    \label{fig:split task1 output}
\end{figure}

\begin{figure}[H]
\centering
    \ContinuedFloat*
    \includegraphics[scale=0.25]{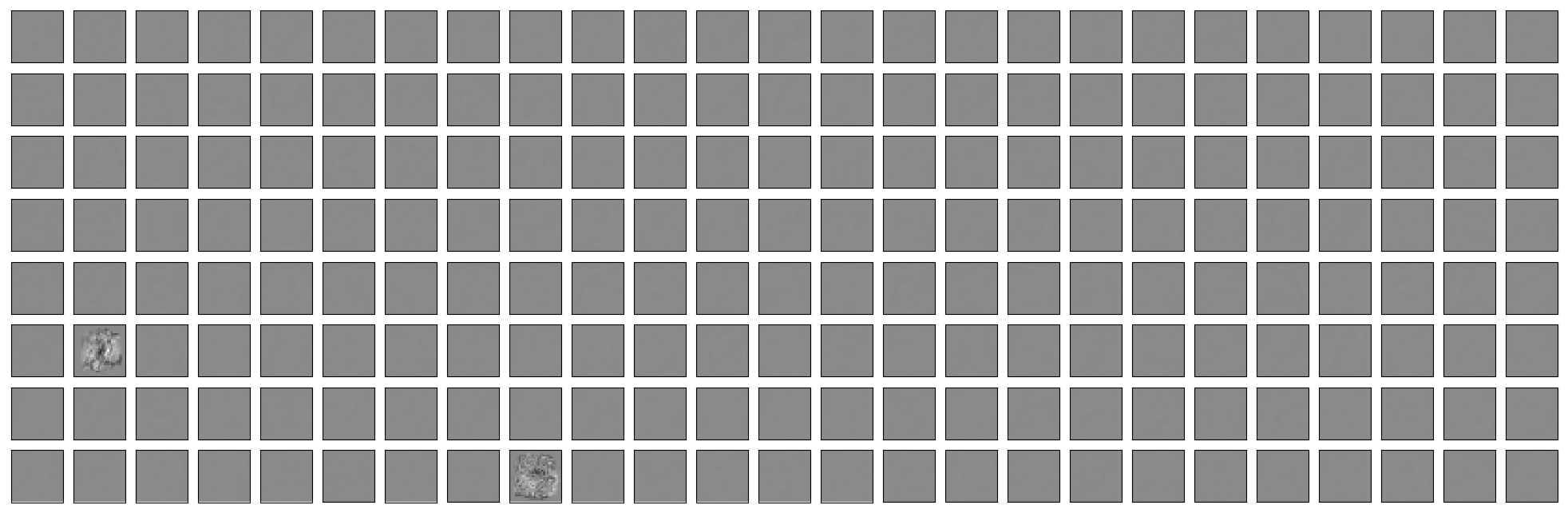}
    \caption{Hidden layer input means after training on task 2}
    \label{fig:split task2 layer0 mean}
\end{figure}

\begin{figure}[H]
\centering
    \ContinuedFloat
    \includegraphics[scale=0.25]{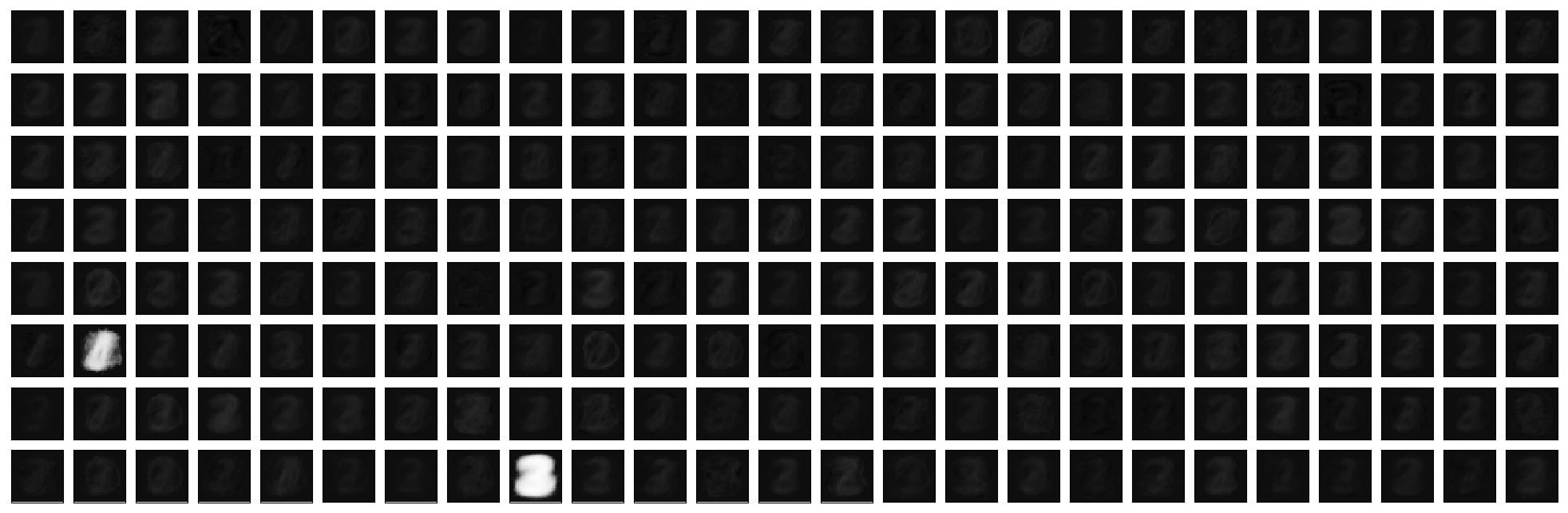}
    \caption{Hidden layer input variances after training on task 2}
    \label{fig:split task2 layer0 var}
\end{figure}

\begin{figure}[H]
\centering
    \ContinuedFloat
    \includegraphics[scale=0.25]{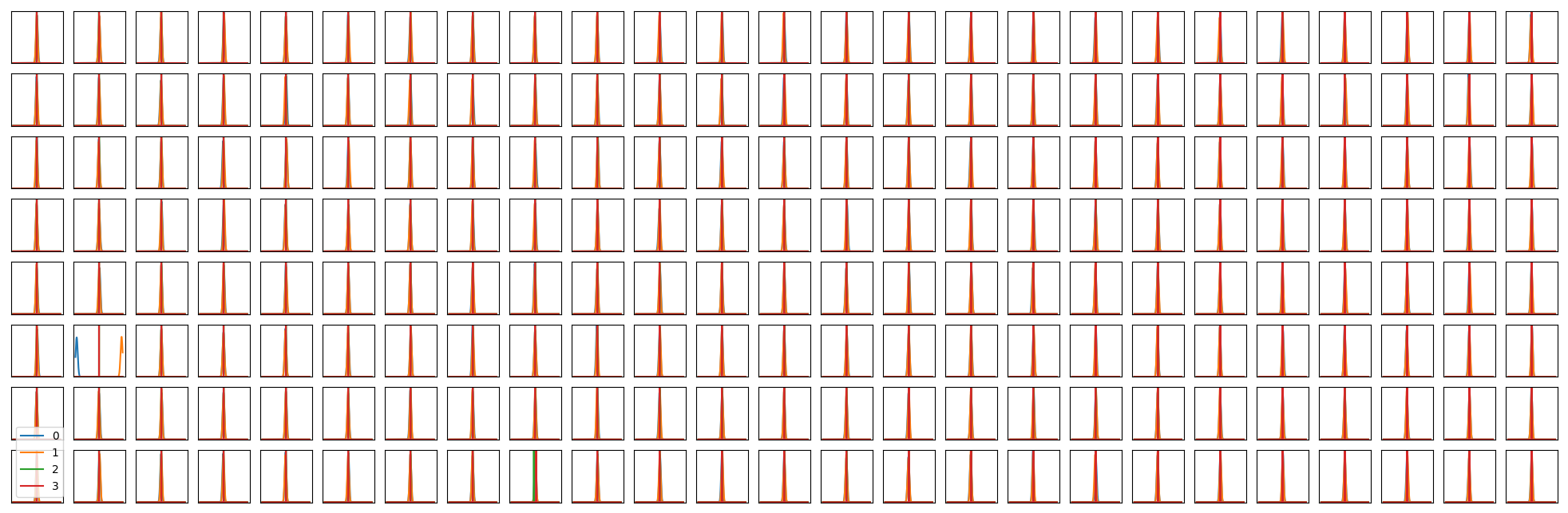}
    \caption{Hidden layer output weights after training on task 2}
    \label{fig:split task2 output}
\end{figure}

\begin{figure}[H]
\centering
    \ContinuedFloat*
    \includegraphics[scale=0.25]{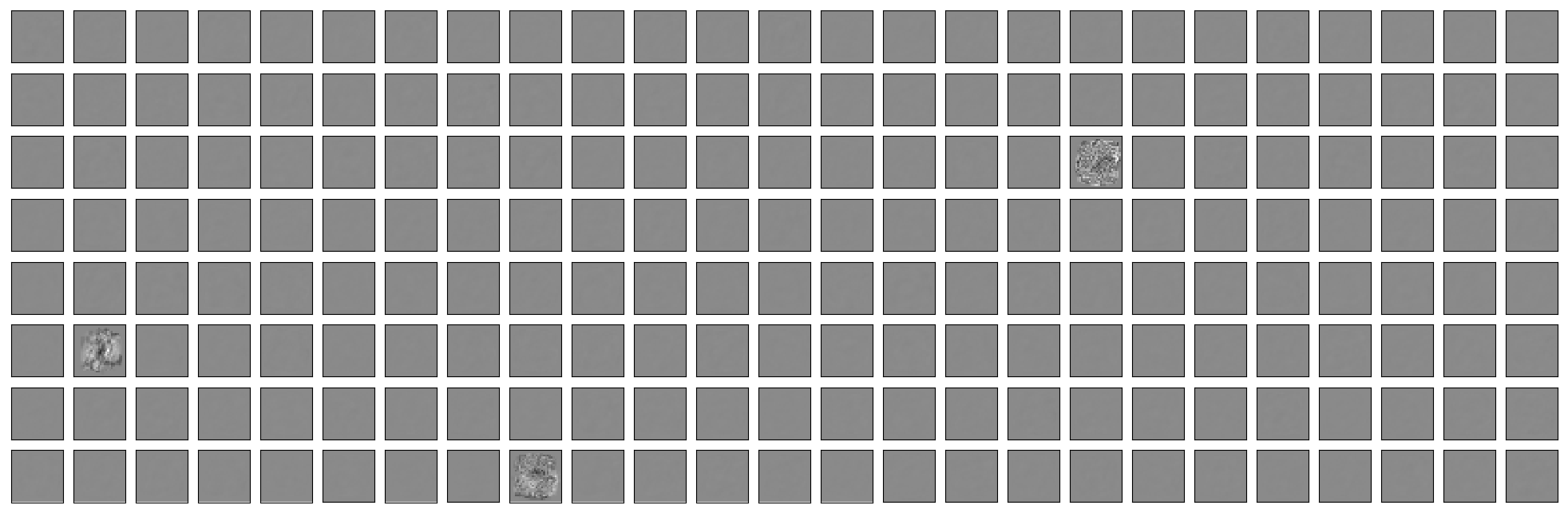}
    \caption{Hidden layer input means after training on task 3}
    \label{fig:split task3 layer0 mean}
\end{figure}

\begin{figure}[H]
\centering
    \ContinuedFloat
    \includegraphics[scale=0.25]{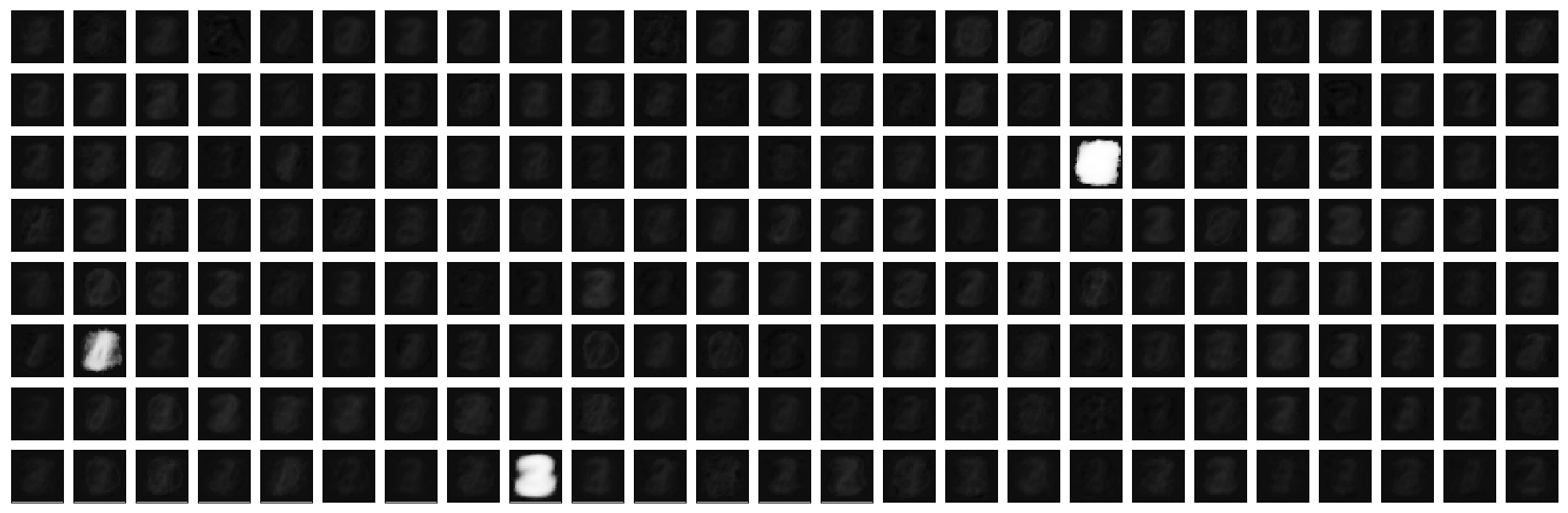}
    \caption{Hidden layer input variances after training on task 3}
    \label{fig:split task3 layer0 var}
\end{figure}

\begin{figure}[H]
\centering
    \ContinuedFloat
    \includegraphics[scale=0.25]{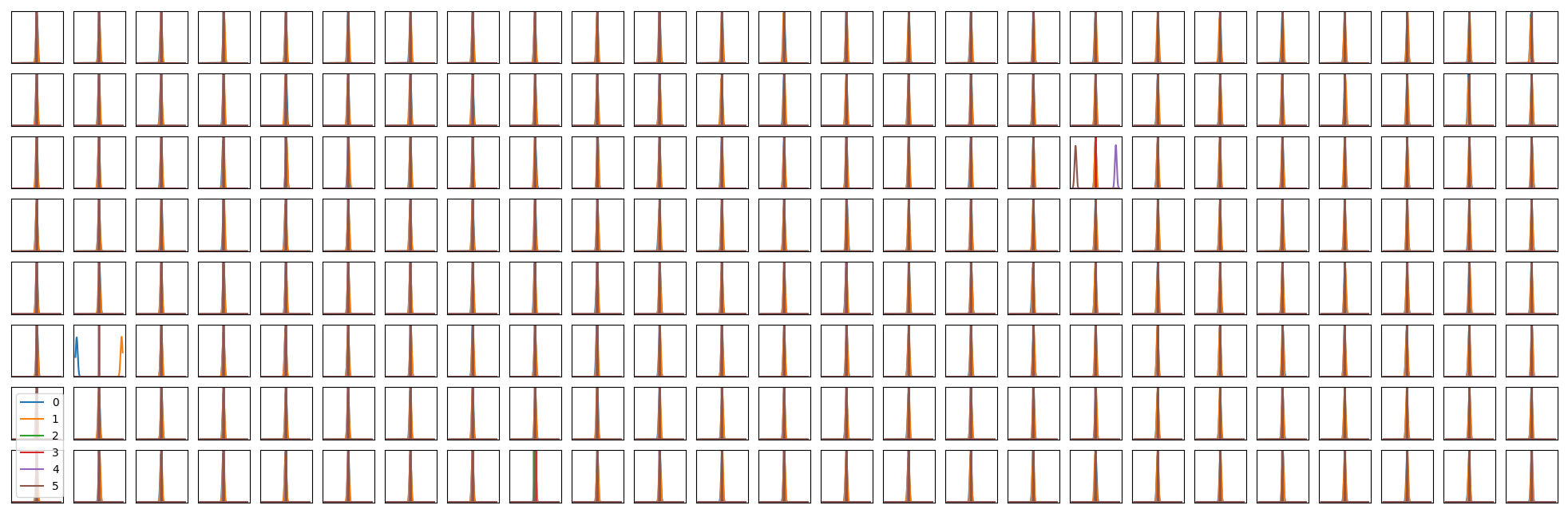}
    \caption{Hidden layer output weights after training on task 3}
    \label{fig:split task3 output}
\end{figure}

\begin{figure}[H]
\centering
    \ContinuedFloat*
    \includegraphics[scale=0.25]{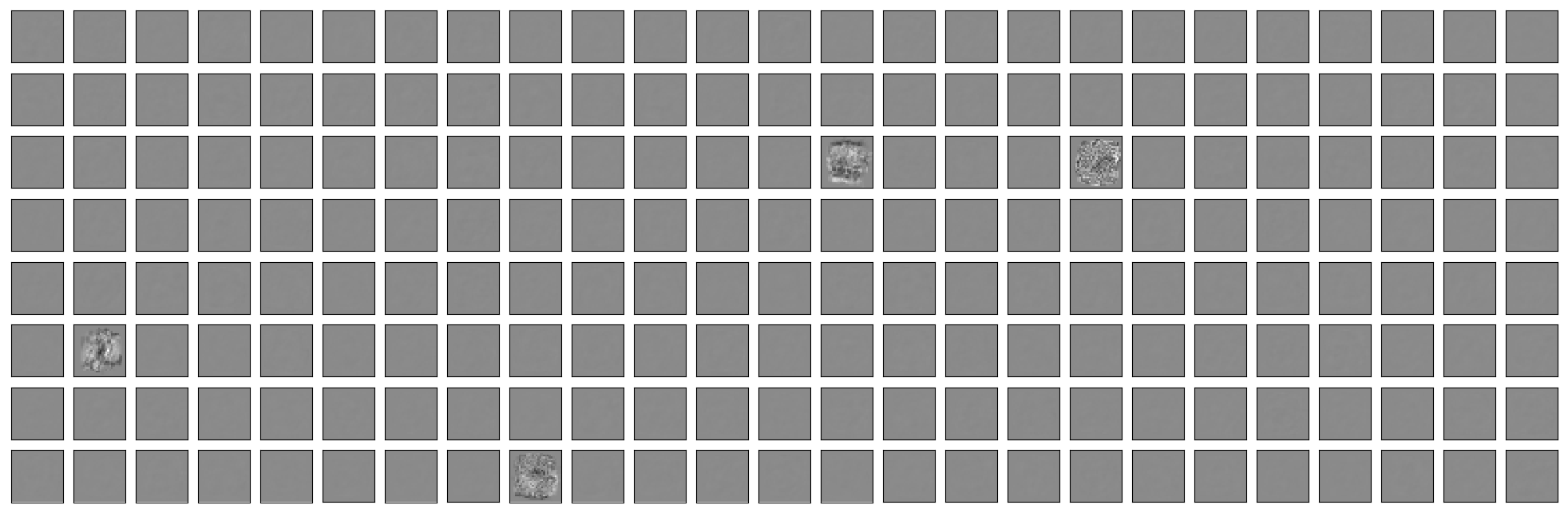}
    \caption{Hidden layer input means after training on task 4}
    \label{fig:split task4 layer0 mean}
\end{figure}

\begin{figure}[H]
\centering
    \ContinuedFloat
    \includegraphics[scale=0.25]{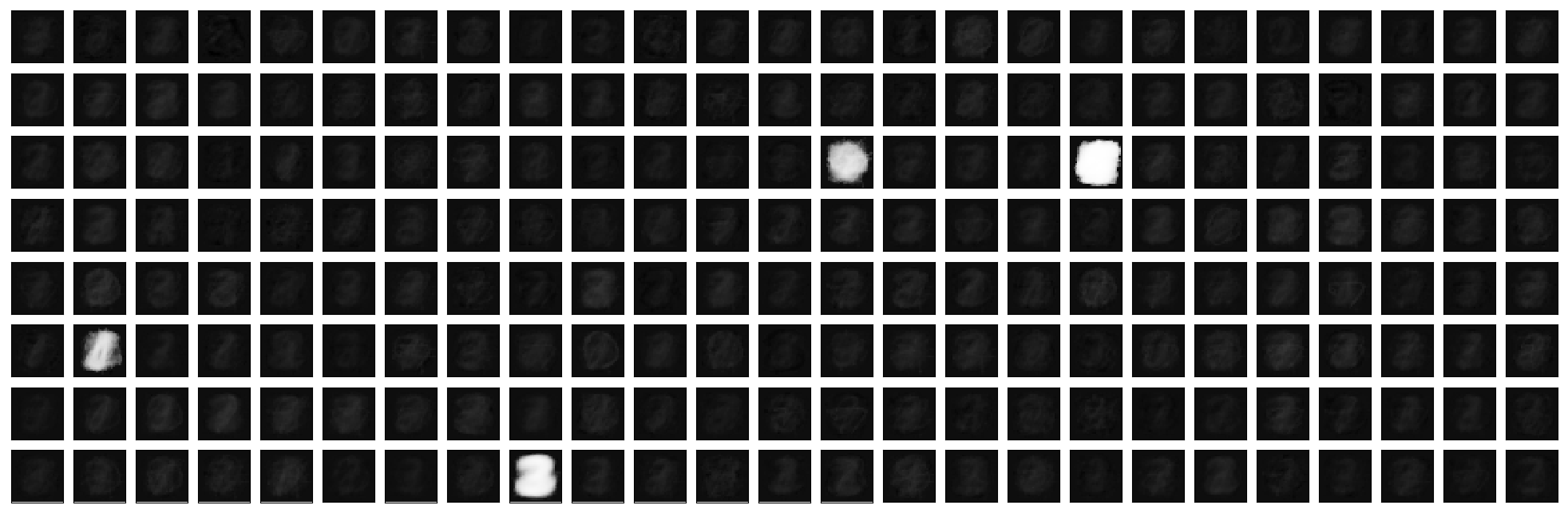}
    \caption{Hidden layer input variances after training on task 4}
    \label{fig:split task4 layer0 var}
\end{figure}

\begin{figure}[H]
\centering
    \ContinuedFloat
    \includegraphics[scale=0.25]{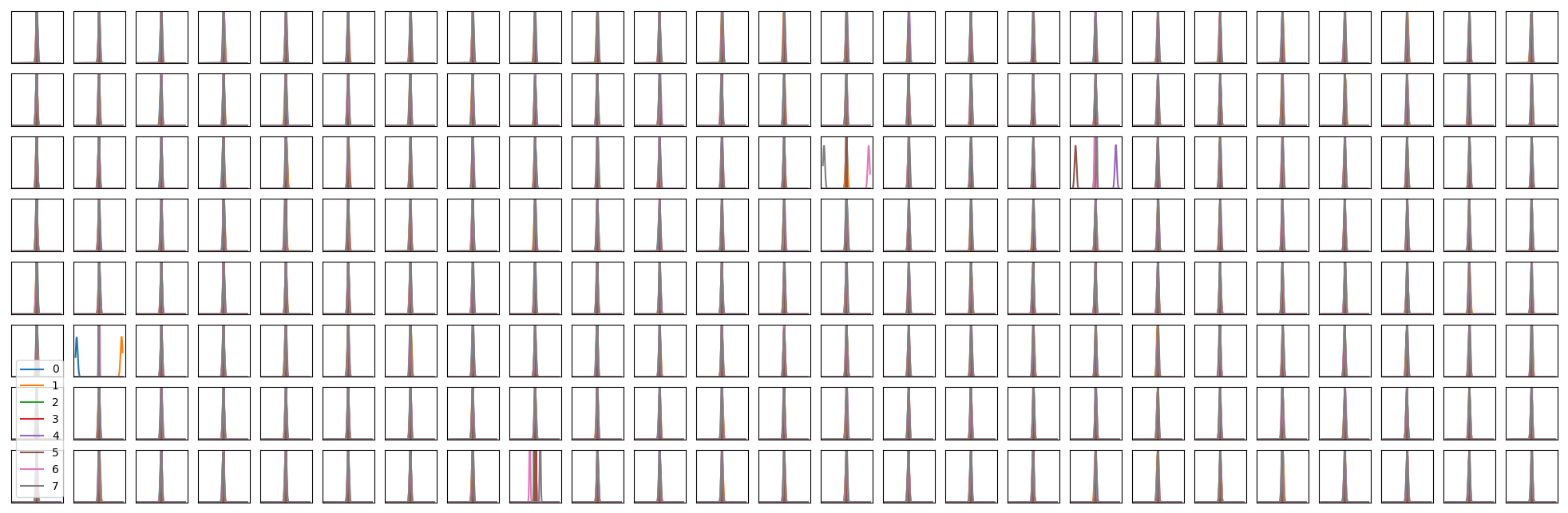}
    \caption{Hidden layer output weights after training on task 4}
    \label{fig:split task4 output}
\end{figure}

\begin{figure}[H]
\centering
    \ContinuedFloat*
    \includegraphics[scale=0.25]{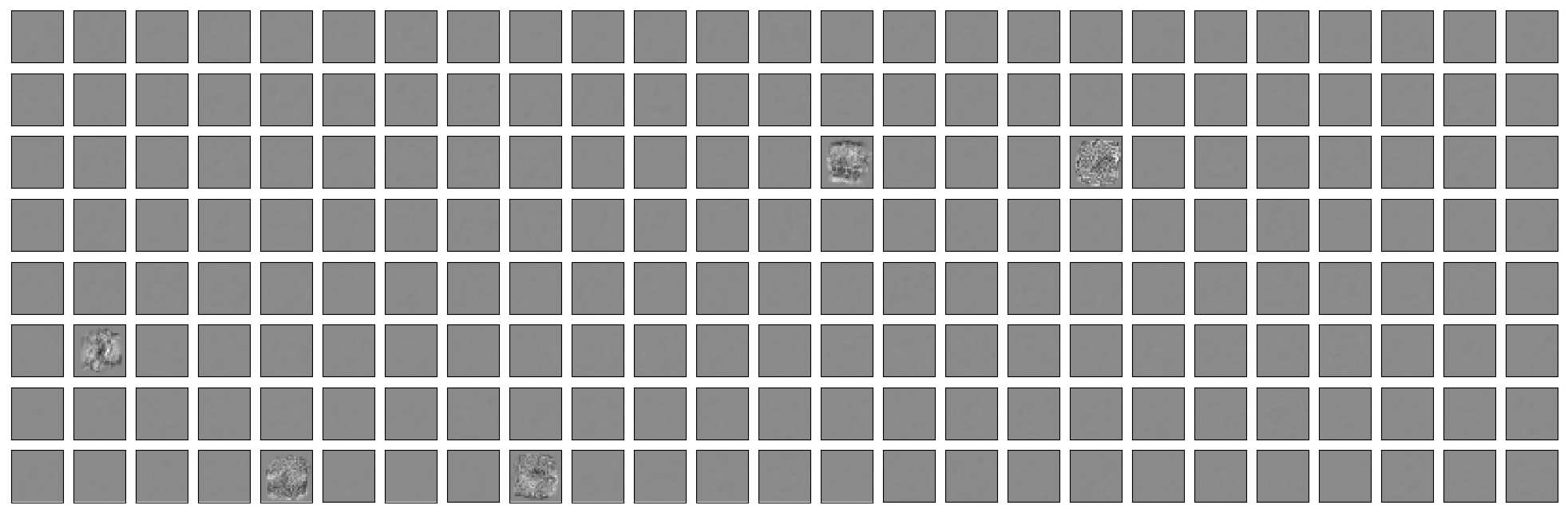}
    \caption{Hidden layer input means after training on task 5}
    \label{fig:split task5 layer0 mean}
\end{figure}

\begin{figure}[H]
\centering
    \ContinuedFloat
    \includegraphics[scale=0.25]{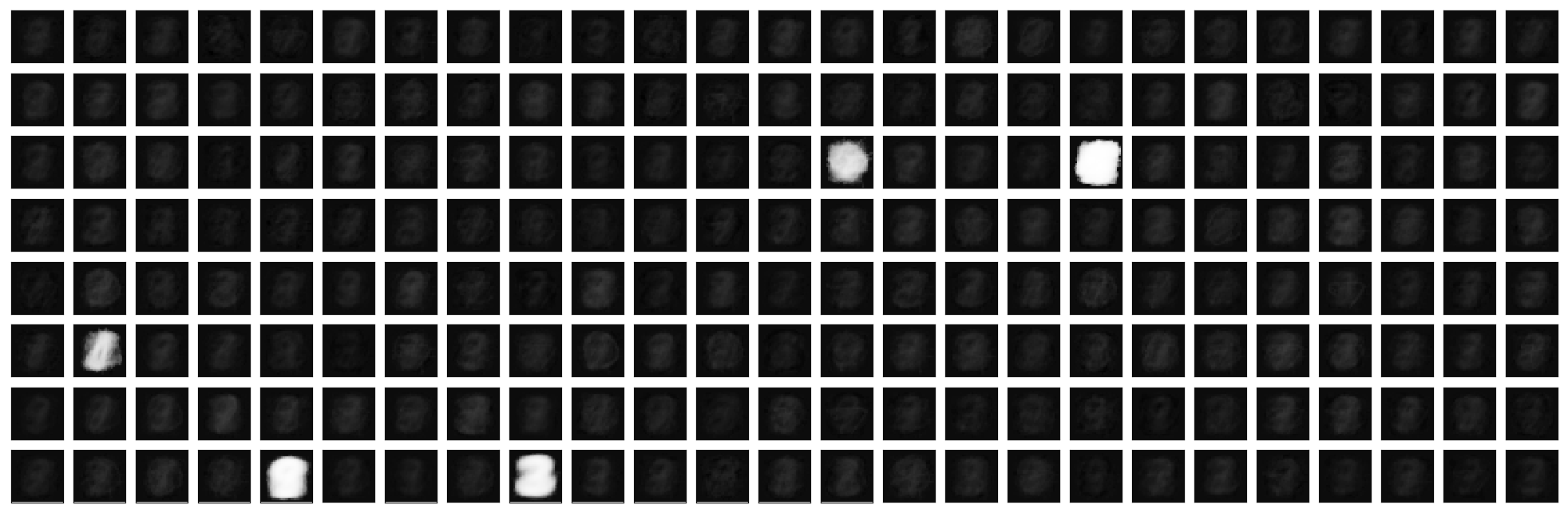}
    \caption{Hidden layer input variances after training on task 5}
    \label{fig:split task5 layer0 var}
\end{figure}

\begin{figure}[H]
\centering
    \ContinuedFloat
    \includegraphics[scale=0.25]{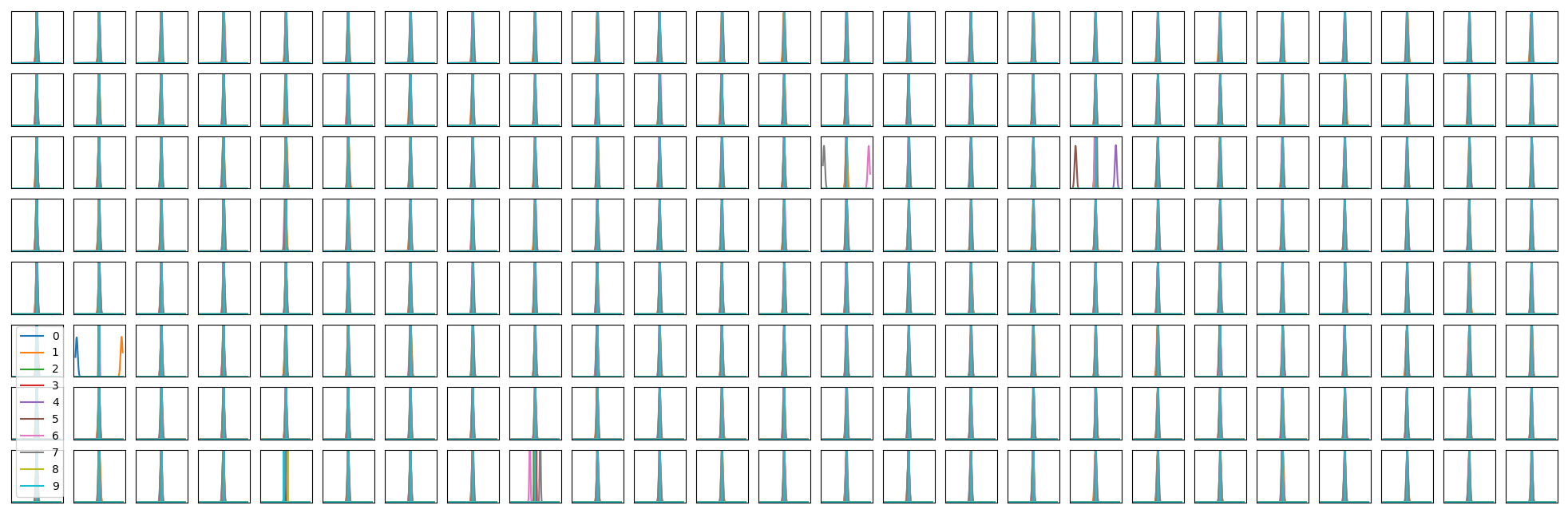}
    \caption{Hidden layer output weights after training on task 5}
    \label{fig:split task5 output}
\end{figure}

\section{} \label{app:permuted}
This appendix has figures showing the means and variances of weights after training without coresets on tasks 1, 5 and 10 on permuted MNIST (MLP with 2 hidden layers, 100 units in each layer, 10 tasks total).

\begin{figure}[H]
\centering
    \ContinuedFloat*
    \includegraphics[scale=0.3]{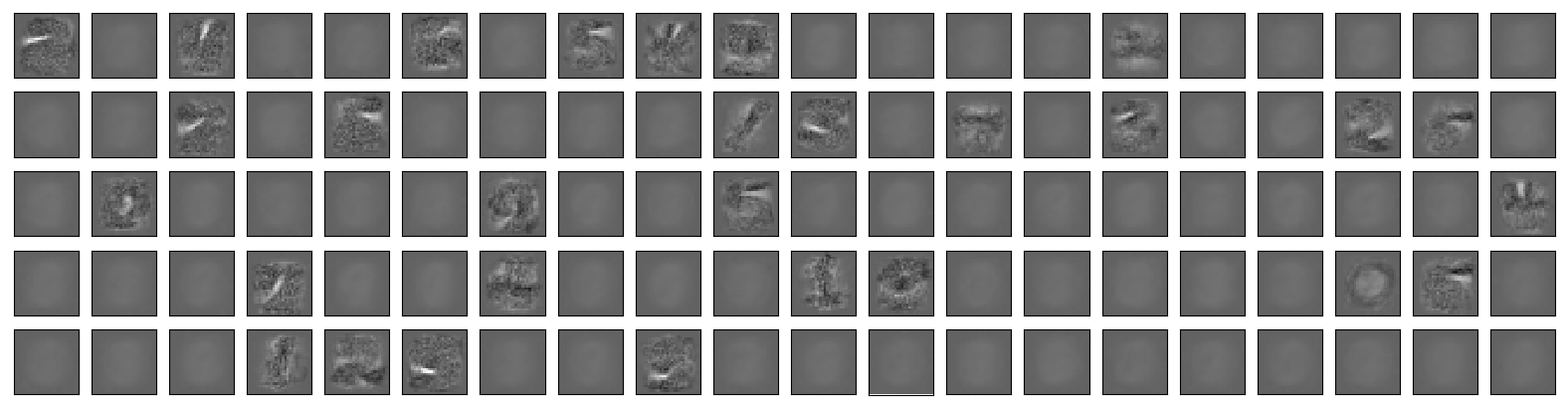}
    \caption{1st hidden layer input means after training on task 1}
    \label{fig:permuted task1 layer0 mean}
\end{figure}

\begin{figure}[H]
\centering
    \ContinuedFloat
    \includegraphics[scale=0.3]{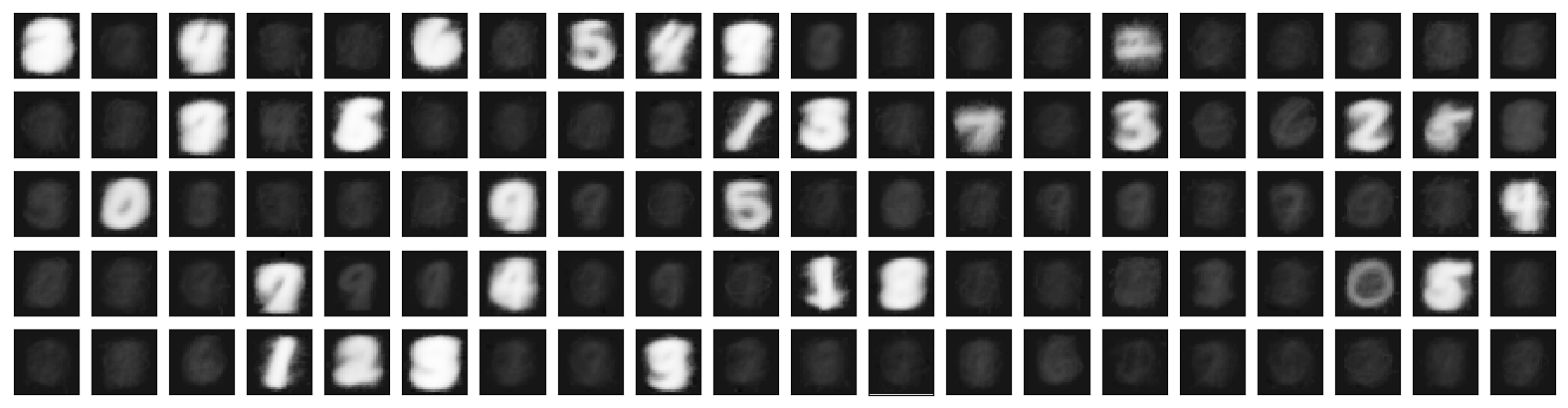}
    \caption{1st hidden layer input variances after training on task 1}
    \label{fig:permuted task1 layer0 var}
\end{figure}

\begin{figure}[H]
\centering
    \ContinuedFloat
    \includegraphics[scale=0.3]{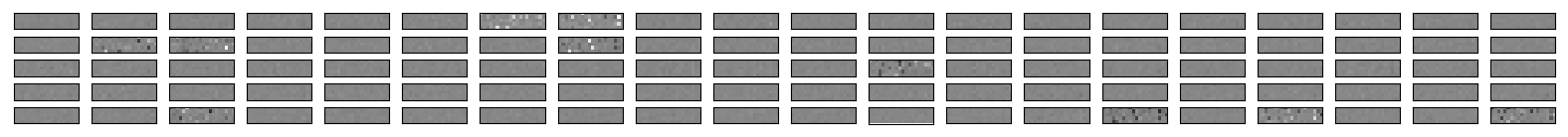}
    \caption{2nd hidden layer input means after training on task 1}
    \label{fig:permuted task1 layer1 mean}
\end{figure}

\begin{figure}[H]
\centering
    \ContinuedFloat
    \includegraphics[scale=0.3]{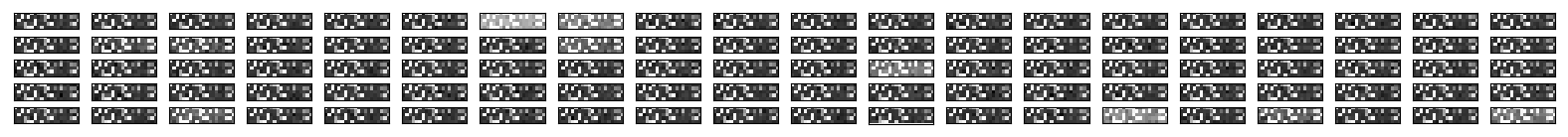}
    \caption{2nd hidden layer input variances after training on task 1}
    \label{fig:permuted task1 layer1 var}
\end{figure}

\begin{figure}[H]
\centering
    \ContinuedFloat
    \includegraphics[scale=0.3]{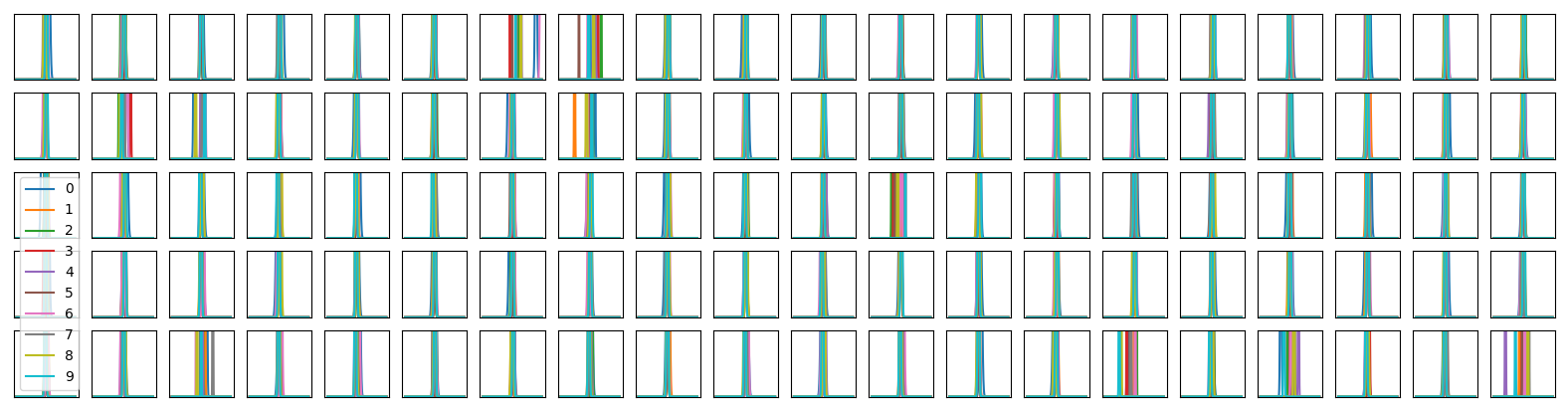}
    \caption{2nd hidden layer output weights after training on task 1 (best viewed in colour)}
    \label{fig:permuted task1 output}
\end{figure}

\begin{figure}[H]
\centering
    \ContinuedFloat*
    \includegraphics[scale=0.3]{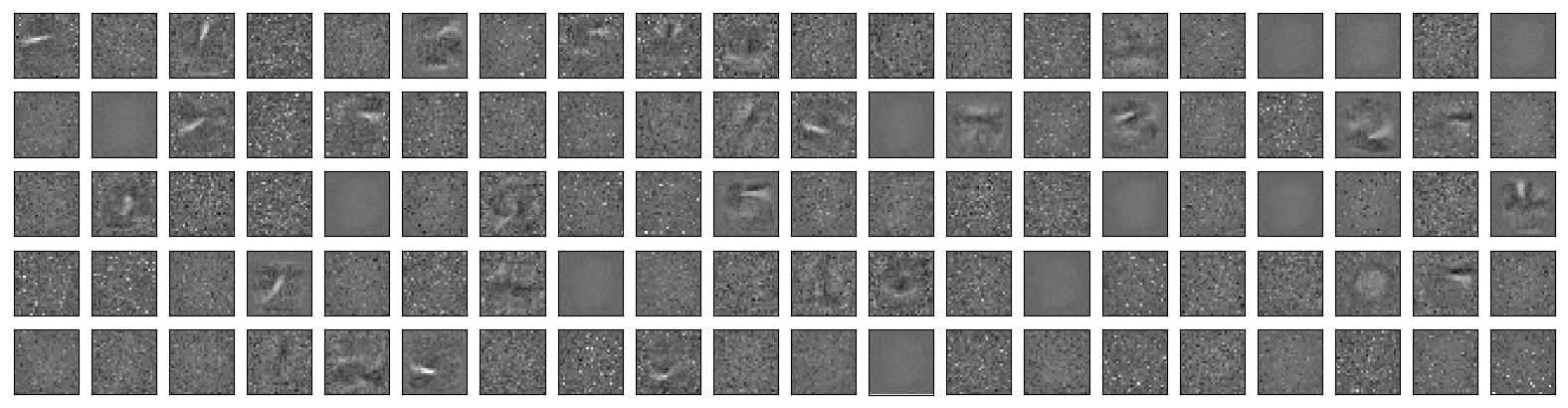}
    \caption{1st hidden layer input means after training on task 5}
    \label{fig:permuted task5 layer0 mean}
\end{figure}

\begin{figure}[H]
\centering
    \ContinuedFloat
    \includegraphics[scale=0.3]{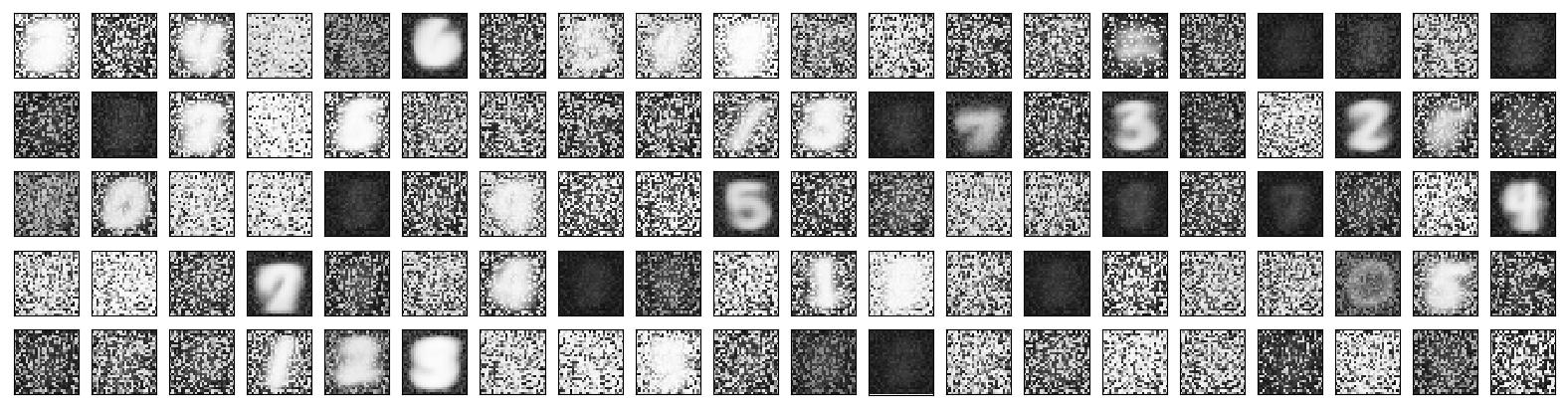}
    \caption{1st hidden layer input variances after training on task 5}
    \label{fig:permuted task5 layer0 var}
\end{figure}

\begin{figure}[H]
\centering
    \ContinuedFloat
    \includegraphics[scale=0.3]{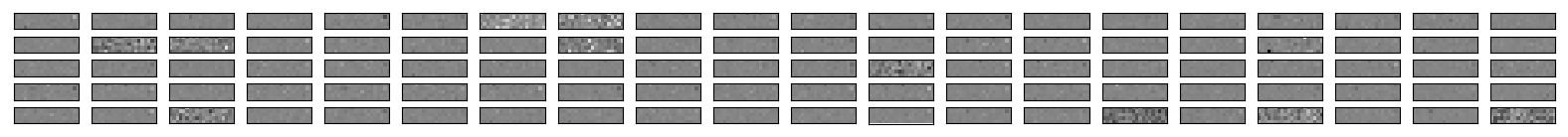}
    \caption{2nd hidden layer input means after training on task 5}
    \label{fig:permuted task5 layer1 mean}
\end{figure}

\begin{figure}[H]
\centering
    \ContinuedFloat
    \includegraphics[scale=0.3]{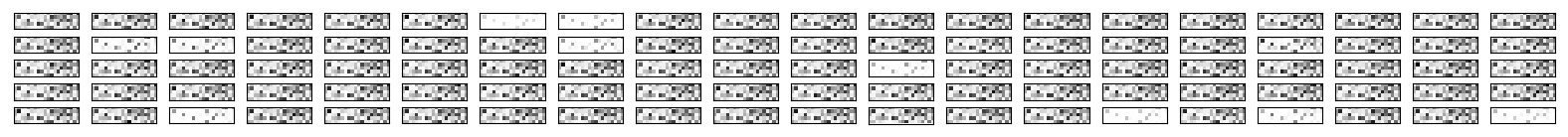}
    \caption{2nd hidden layer input variances after training on task 5}
    \label{fig:permuted task5 layer1 var}
\end{figure}

\begin{figure}[H]
\centering
    \ContinuedFloat
    \includegraphics[scale=0.3]{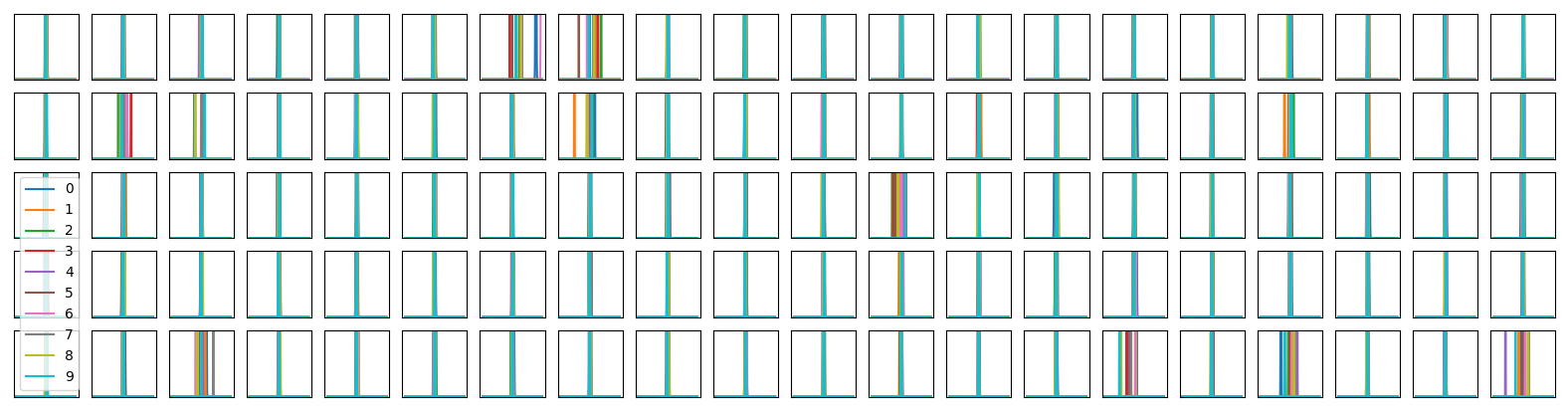}
    \caption{2nd hidden layer output weights after training on task 5 (best viewed in colour)}
    \label{fig:permuted task5 output}
\end{figure}

\begin{figure}[H]
\centering
    \ContinuedFloat*
    \includegraphics[scale=0.3]{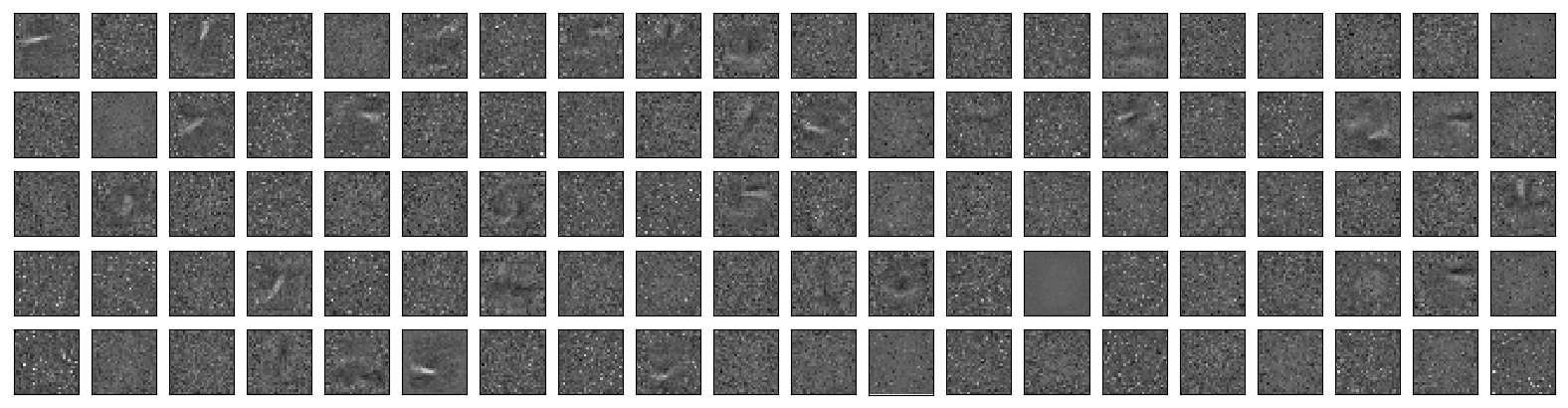}
    \caption{1st hidden layer input means after training on task 10}
    \label{fig:permuted task10 layer0 mean}
\end{figure}

\begin{figure}[H]
\centering
    \ContinuedFloat
    \includegraphics[scale=0.3]{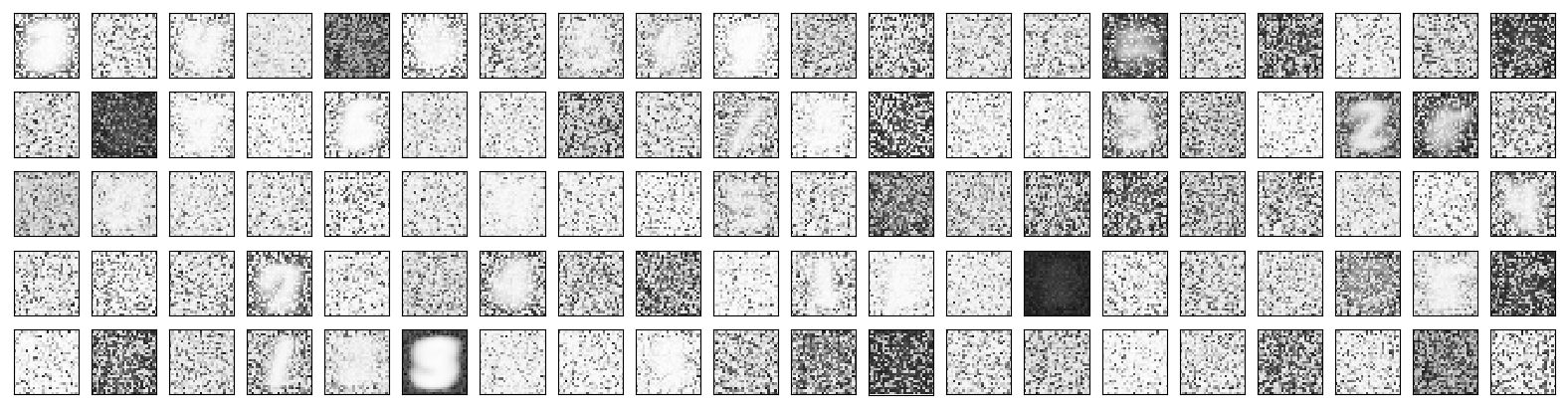}
    \caption{1st hidden layer input variances after training on task 10}
    \label{fig:permuted task10 layer0 var}
\end{figure}

\begin{figure}[H]
\centering
    \ContinuedFloat
    \includegraphics[scale=0.3]{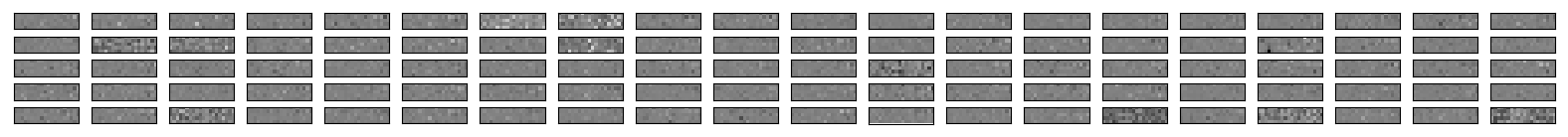}
    \caption{2nd hidden layer input means after training on task 10}
    \label{fig:permuted task10 layer1 mean}
\end{figure}

\begin{figure}[H]
\centering
    \ContinuedFloat
    \includegraphics[scale=0.3]{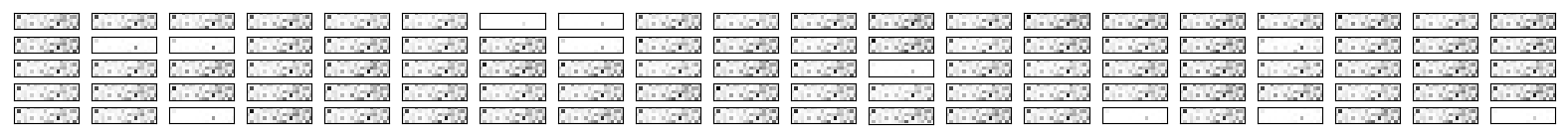}
    \caption{2nd hidden layer input variances after training on task 10}
    \label{fig:permuted task10 layer1 var}
\end{figure}

\begin{figure}[H]
\centering
    \ContinuedFloat
    \includegraphics[scale=0.3]{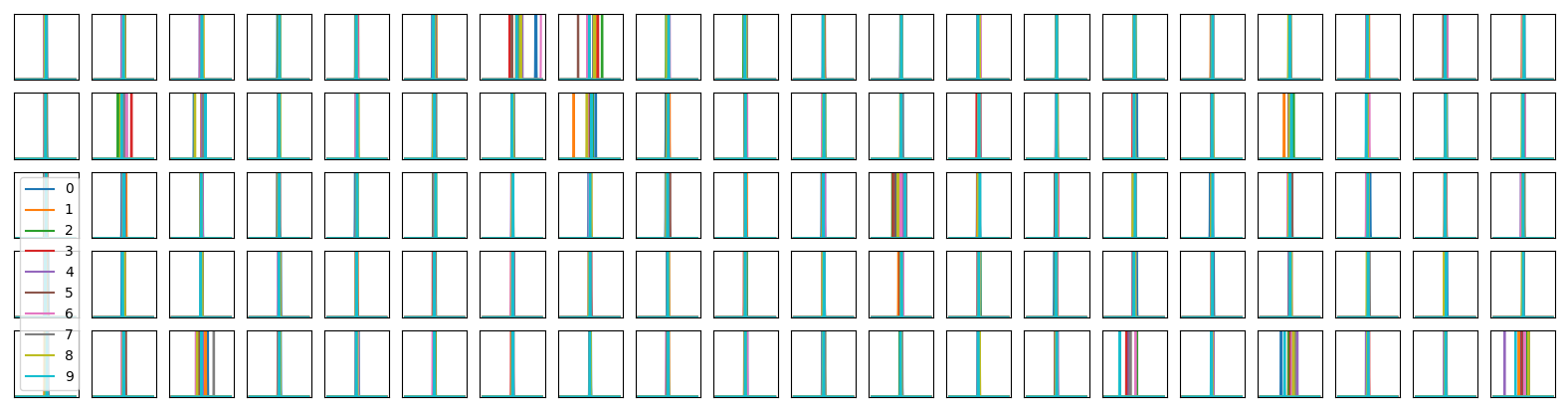}
    \caption{2nd hidden layer output weights after training on task 10 (best viewed in colour)}
    \label{fig:permuted task10 output}
\end{figure}

\end{document}